\documentclass{article}

\usepackage[preprint,nonatbib]{neurips_2024}

\usepackage{graphicx}
\usepackage{xcolor}
\usepackage{amsmath}
\usepackage{tcolorbox}

\usepackage[utf8]{inputenc} 
\usepackage[T1]{fontenc}    
\usepackage{hyperref}       
\usepackage{url}            
\usepackage{booktabs}       
\usepackage{amsfonts}       
\usepackage{nicefrac}       
\usepackage{microtype}      
\usepackage{xcolor}         
\usepackage{caption}
\usepackage{subcaption}
\usepackage{bm}

\title{Physics of Skill Learning}

\author{%
  Ziming Liu$^{1,2}$\thanks{zmliu@mit.edu}\quad Yizhou Liu$^{1}$\quad Eric J. Michaud$^{1,2}$\quad Jeff Gore$^{1}$\quad Max Tegmark$^{1,2}$ \\
  $^1$ Department of Physics, Massachusetts Institute of Technology \\
  $^2$ The NSF Institute for Artificial Intelligence and Fundamental Interactions
}

\begin{document}

\maketitle

\begin{abstract}
   We aim to understand \textit{physics of skill learning}, i.e., how skills are learned in neural networks during training. We start by observing the \textbf{Domino effect}, i.e., skills are learned sequentially, and notably, some skills kick off learning right after others complete learning, similar to the sequential fall of domino cards. To understand the Domino effect and relevant behaviors of skill learning, we take physicists' approach of abstraction and simplification. We propose three models with varying complexities -- the \textbf{Geometry model}, the \textbf{Resource model}, and the \textbf{Domino model}, trading between reality and simplicity. The Domino effect can be reproduced in the Geometry model, whose resource interpretation inspires the Resource model, which can be further simplified to the Domino model. These models present different levels of abstraction and simplification; each is useful to study some aspects of skill learning. The Geometry model provides interesting insights into neural scaling laws and optimizers; the Resource model sheds light on the learning dynamics of compositional tasks; the Domino model reveals the benefits of modularity. These models are not only conceptually interesting -- e.g., we show how Chinchilla scaling laws can emerge from the Geometry model, but also are useful in practice by inspiring algorithmic development -- e.g., we show how simple algorithmic changes, motivated by these toy models, can speed up the training of deep learning models.
\end{abstract}

\begin{figure}[ht]
    \centering
    \includegraphics[width=1.0\linewidth, trim={4cm 0 3cm 0}]{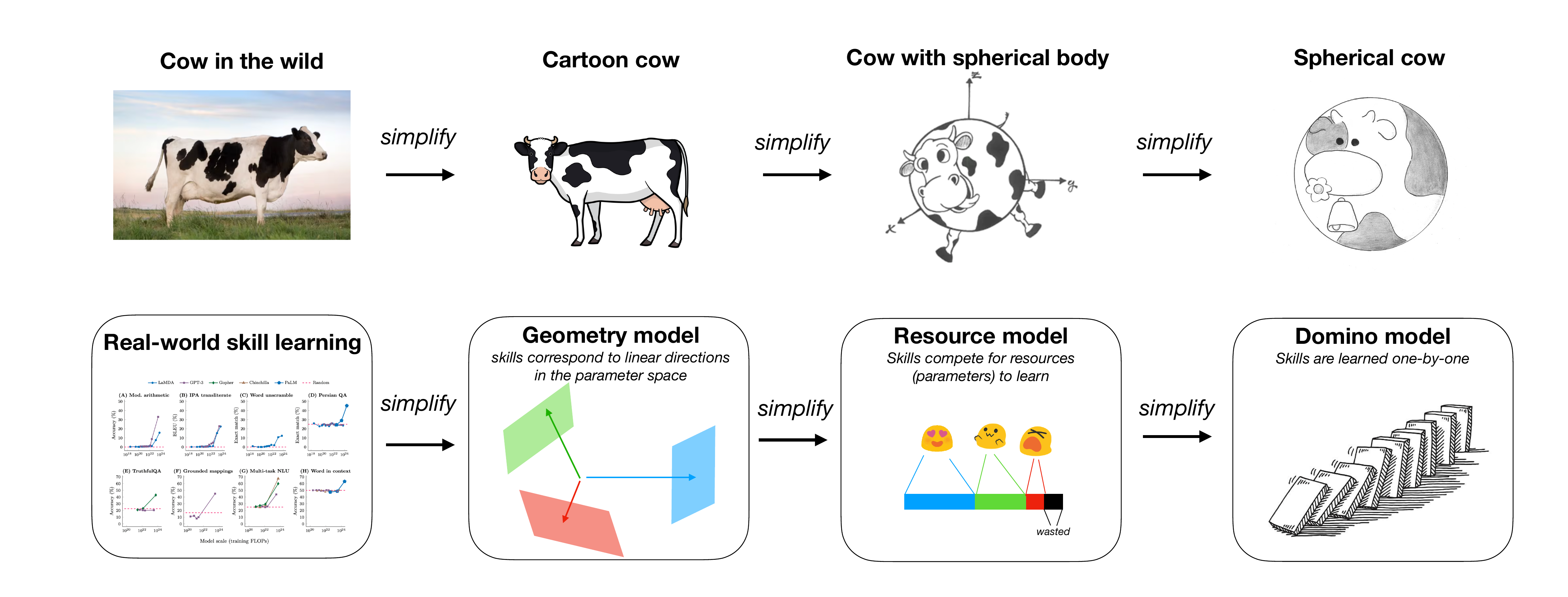}
    \caption{Physicists are famous for making up (sometimes overly) simplistic models. 
    We take the same spirit to understand skill learning. We propose three models trading off between reality and simplicity: the Geometry model, the Resource model, and the Domino model.}
    \label{fig:philosophy}
\end{figure}

\section{Introduction}

\begin{figure}[t]
    \centering
    \includegraphics[width=0.9\linewidth]{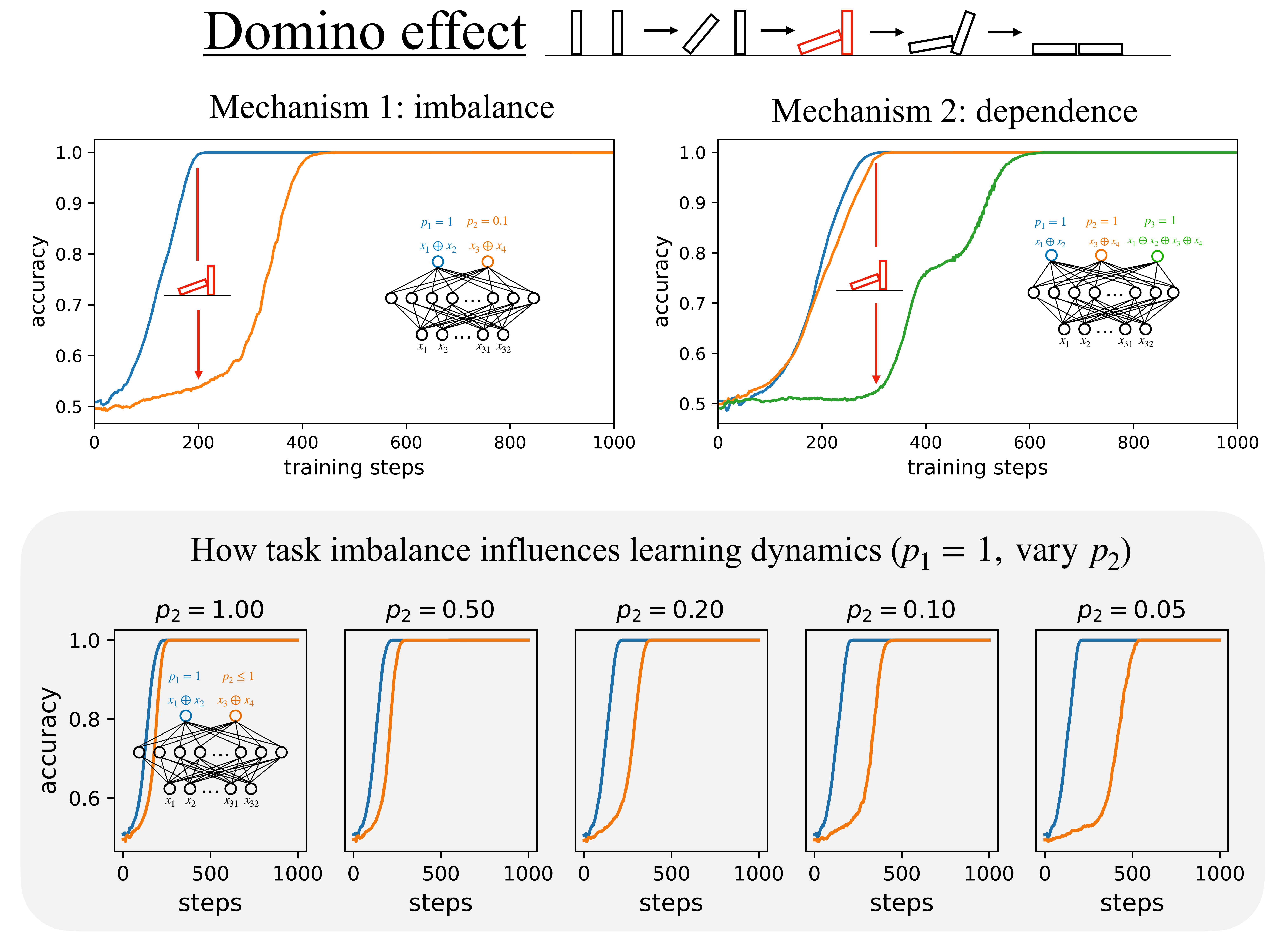}
    \caption{The Domino effect (sequential learning of tasks) occurs for sparse parity learning. Top: The Domino effect can be attributed to task imbalance (left) or compositional dependency (right). Bottom: how task imbalance influences learning dynamics.}
    \label{fig:sp_two_mechs}
\end{figure}

Language models are demonstrating impressive skills in, e.g., coding and mathematics. Many tasks, including language modeling, are complex composite tasks that can be decomposed into many atomic skills~\cite{arora2023theory, chen2024skill, michaud2024quantization, wei2022emergent,ortiz2024task,ilharco2022editing}. The learning dynamics of skills appear to be complex and intriguing: Throughout training, a skill can be completely learned, partially learned, or not learned at all. Even for learned skills, they could display quite diverse learning curves, including sudden jumps (grokking), gradual improvements, or non-monotonic oscillations. Despite the diverse phenomenology observed in real-world experiments, our intuitive understanding of them is quite limited. 
Intuitive understanding, or physics-like understanding, has the potential to bridge between theory (mathematics-like understanding) and experiments (engineering-like understanding). 

To gain some intuition about skill learning, we take physicists' approach of abstraction and simplification (see illustration in Figure~\ref{fig:philosophy}): when trying to understand a cow in the wild, physicists would make assumptions to simplify the subject matter. It is science but also art to determine the appropriate level of abstraction and simplification. As Einstein famously put it, ``Everything should be made as simple as possible, but not simpler.'' In the same philosophy, we will propose three models trading off between reality and simplicity -- the  \textit{Geometry model}, the \textit{Resource model} and the \textit{Domino model}. Each of these models is able to capture some realistic aspects of rich skill dynamics. Note that people use similar words for skills, e.g., tasks, quanta, abilities, etc. In this paper, we stick to the terms ``skill'' and ``task'' and use them interchangeably (pedagogically, ``a skill = the ability to perform a task''). 

As a motivation, we start by making an observation called the \textit{Domino effect}, which shows that skills tend to learn sequentially, and notably, some skills start to learn right after other skills finish learning. For example, when we train two independent sparse parity tasks (with frequencies $p_1=1$ and $p_2=0.1$, shown in Figure~\ref{fig:sp_two_mechs} left) on a two-layer MLP using the Adam optimizer, the second task starts to progress rapidly only after the first task finishes. Quantitatively, learning task 2 only takes two more times (instead of $p_1/p_2=10$ times that one would reasonably expect since the gradient signals differ by 10 times). In a more complicated setup (shown in Figure~\ref{fig:sp_two_mechs} right), compositional task dependency can also lead to the Domino effect. It is thus very intriguing to understand the mechanisms underneath the Domino effect. Although our ambitious goal is to understand task dynamics in general, the Domino effect serves as the starting point.

\begin{figure}[t]
    \centering
    \includegraphics[width=1.0\linewidth]{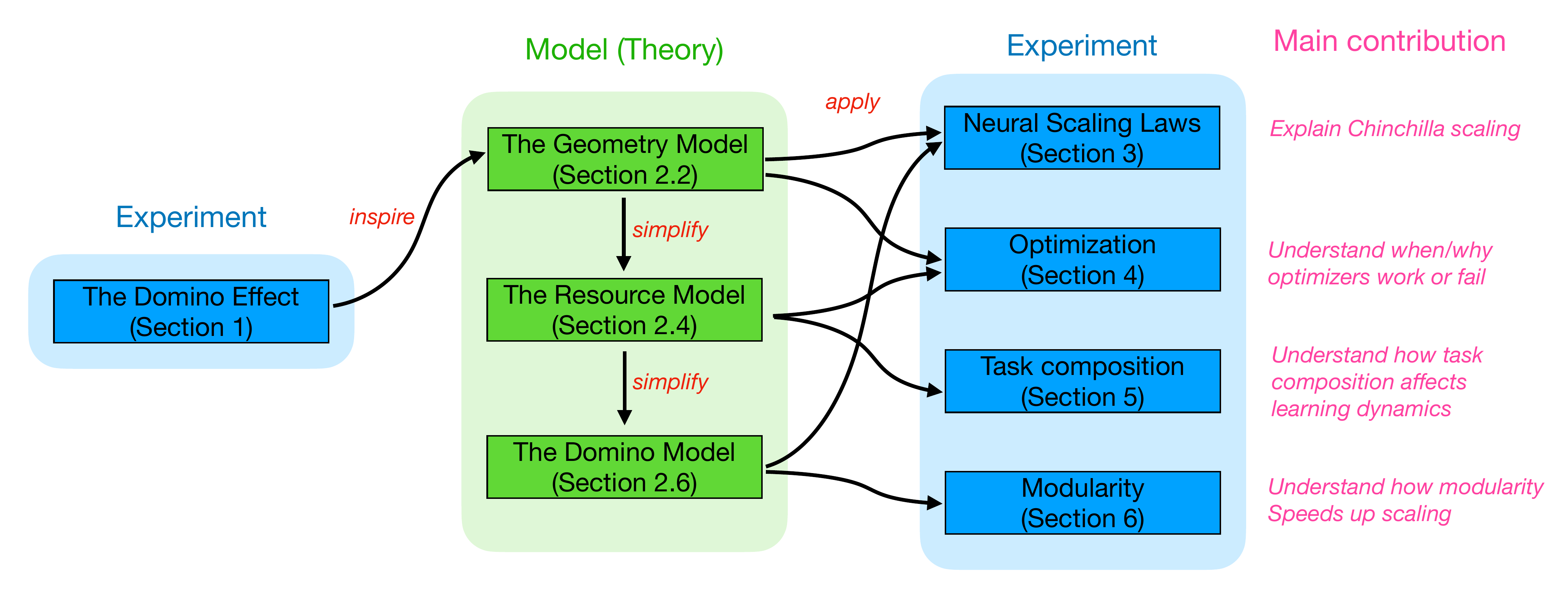}
    \caption{Organization. Physics-like theories (models) are inspired by experiments and contribute to guiding the design of new experiments.}
    \label{fig:organization}
\end{figure}

We first propose the \textit{Geometry model} in which the Domino effect can occur. When attempting to understand the phenomenology of the Geometry model, a resource interpretation is developed and is then translated to the \textit{Resource model}. The Resource model elegantly describes how skills ``fight'' for constrained ``resources''. When tasks have strong hierarchical structures, the Resource model can be further simplified to \textit{the Domino model}, in which skills are learned in a strict sequential order.  Although the initial motivation of these models was to understand the Domino effect, their applications can go much broader than their initial scopes. The Geometry model provides insights into neural scaling laws and optimizers; the Resource model sheds light on the learning dynamics of compositional tasks; the Domino model provides an analysis framework for modularity. The organization of the paper is illustrated in Figure~\ref{fig:organization}: In section~\ref{sec:models}, we propose models of varying complexities to explain the Domino effect. These models are not only conceptually interesting, but also practically useful, with their applications to neural scaling laws in section \ref{sec:nsl}, optimization in section~\ref{sec:optimization}, task compositionality in section~\ref{sec:dependent}, and modularity in section~\ref{sec:modularity}. Codes are available at \url{https://github.com/KindXiaoming/physics_of_skill_learning}.

\section{Three Models of Skill Learning: From cows in the wild to spherical cows in a vacuum}\label{sec:models}

We aim to propose models that manifest the Domino effect shown in Figure~\ref{fig:sp_two_mechs}. We start by introducing the philosophy in Section~\ref{subsec:philosophy}. The philosophy leads to the Geometry model in Section~\ref{subsec:mechanistic}. In Section~\ref{subsec:overparam}, we analyze the phenomenological behavior of the Geometry model in the overparametrized regime, observing the Domino effect. In Section~\ref{subsec:effective}, we give a resource interpretation to the Geometry model, resulting in the (simpler) Resource model. In section~\ref{subsec:underparam}, we study both the Geometry model and the Resource model in the underparametrized regime. In Section~\ref{subsec:domino}, we further simplify the Resource model to the Domino model, by assuming a strong hierarchy of tasks. 

\subsection{Philosophy: Coarse-graining your models}\label{subsec:philosophy}

Einstein has famously stated that ``models should be made simple, but not simpler''. This means that we want to make models simple enough in order for intuition to work, but also expressive enough to be relevant to the phenomenon we are interested in. For example, to model the behavior of ideal gas, it suffices to have the ideal gas law $PV = NRT$ which states that only four macroscopic variables are important: pressure $P$, volume $V$, temperature $T$, and the number of atoms $NR$. It is unnecessary to keep track of the positions and velocities of all gas atoms. This is a common philosophy in physics, where we want to ``coarse-grain'' many irrelevant microscopic details to relevant macroscopic variables, reducing the model complexity to a great extent. 

How can we understand learning dynamics (determined by loss landscape and optimizers)? The full loss landscape depends on all model parameters, i.e., $\ell = \ell_p(\bm{\theta})$ (subscript $p$ stands for parameters), where $\bm{\theta}$ usually contains more than millions or even billions of parameters. We want to ``coarse-grain'' these parameters to microscopic parameters which describe the network's skills, e.g., translating between English to Chinese, the ability to do multiplication, the ability to do induction, etc. To reduce parameters to skills, we assume there exists a compressor $C$ such that $\bm{s}\equiv C(\bm{\theta})$ describes the abilities of the network, and there exists a loss function $\ell_s$ acting on $\bm{s}$ such that  $\ell_s(\bm{s})\approx \ell_p(\bm{\theta})$.  For simplicity, we denote $\ell_s$ as $\ell$ below.

To get the first model, we need two more assumptions: (1) We first assume that skills are independently contributing to the overall loss such that $\ell=\sum_{i=1}^{n_{\rm task}}p_i\ell (s_i)$ where $p_i$ is the weight/frequencies of the $i^{\rm th}$ skill/task, and $\ell(s_i)$ is the loss induced by the $i^{\rm th}$ skill when the skill level is $s_i$. (2) The compression is linear with respect to $\bm{\theta}$, i.e, there exists a linear direction $\bm{t}_i$ for the $i^{\rm th}$ task such that $s_i=(\bm{\theta}-\bm{\theta}_0)\cdot \bm{t}_i$, where $\bm{\theta}_0$ is the initialized model. This leads to the Geometry model we introduce in the next subsection. 

\subsection{Introducing The Geometry Model}\label{subsec:mechanistic}

\begin{center}
\begin{tcolorbox}[colframe=black, boxrule=1pt, colback=white, width=0.95\textwidth]
{\bf Geometry Model}: Suppose that a network with parameters $\bm{\theta}\in\mathbb{R}^{n_{\rm dim}}$ is tasked with $n_{\rm task}$ independent tasks. Tasks have frequencies (importance weight) $(p_1,p_2,\cdots, p_{n_{\rm task}})$ where $p_1\geq p_2\geq\cdots\geq p_{n_{\rm task}} > 0$. We assume that tasks are linearly represented in the model parameter space -- the $i^{\rm th}$ task is associated with a task vector $\bm{t}_i\in\mathbb{R}^{n_{\rm dim}}$ in the parameter space. Starting from the initialization $\bm{\theta}_0$, the model with parameter $\bm{\theta}$ has the skill level for the $i^{\rm th}$ task $s_i \equiv (\bm{\theta}-\bm{\theta}_0)\cdot \bm{t}_i$. The loss is $\ell = \sum_{i=1}^{n_{\rm task}}p_i\mathcal{L}(s_i)$ or $\ell = \sum_{i=1}^{n_{\rm task}}\tilde{p}_i\mathcal{L}(s_i)$ where $\tilde{p}$ is the empirical frequency when finite batch effect is of interest, and $\mathcal{L}$ is chosen to be $\mathcal{L}(s)=(1-s)^2$ (regression) or $\mathcal{L}(s)=-{\rm log}(\frac{1}{1+e^{-s}})$ (classification). The loss $\ell$ is then minimized using an optimizer chosen by the user, resulting in skill dynamics $s_i(t)$ and loss evolution $\ell (t)$.
\end{tcolorbox}
\end{center}

There are two assumptions made by the Geometry model: (1) {\bf independence}. Tasks are independent, so the total loss is a weighted average of all tasks, weighted by their frequencies. (2) {\bf linear representation}.  Tasks are linearly represented by the model parameters so that skill levels are determined solely by their projections onto corresponding task vectors. The independence assumption may not be fully realistic, but is usually adopted due to its simplicity (e.g., in ~\cite{michaud2024quantization}), so let us roll with it for now; we will also aim to remove this assumption in Section~\ref{sec:dependent} when we study the compositional structure of tasks. The linear representation assumption is not too unrealistic since it is aligned with recent observations on task arithmetic~\cite{ortiz2024task,ilharco2022editing,todd2023function}. The Geometry model only abstracts away some details (task structure, and how the model represents tasks) while leaving other low-level details (e.g., optimization) intact. Below we can add more specifications to the Geometry model for simulation.



{\bf Choice of task vectors} To model independent skills, we assume task vectors are randomly drawn from a Gaussian distribution $t_i\sim \mathcal{N}(0,\frac{1}{n_{\rm dim}}I)$. We call a model overparametrized if $n_{\rm task}\leq n_{\rm dim}$ but otherwise underparametrized. In the overparametrized case, we further simplify the setup by orthogonalizing these random vectors so that task vectors are orthogonal. This is not an essential assumption but it further simplifies the mental picture (progressing along one task vector does not change the overlaps with other task vectors). In the under-parameterized case, it is impossible to orthogonalize all task vectors hence skill correlation (or superposition) becomes inevitable, which is one of the most intriguing phenomena in today's large language models. 

{\bf Power-law distribution} Following~\cite{michaud2024quantization}, we usually choose a Zipfian-like (power-law) distribution, i.e., $p_i = i^{-\alpha}/\sum_{n=1}^{n_{\rm task}} n^{-\alpha} $, where a larger $\alpha$ dictates a more heavy-tailed distribution. 


\subsection{The Overparametrized regime: $n_{\rm task}\leq n_{\rm dim}$}\label{subsec:overparam}


\begin{figure}[ht]
\centering
\begin{subfigure}{.65\textwidth}
  \centering
  \includegraphics[width=1.0\linewidth]{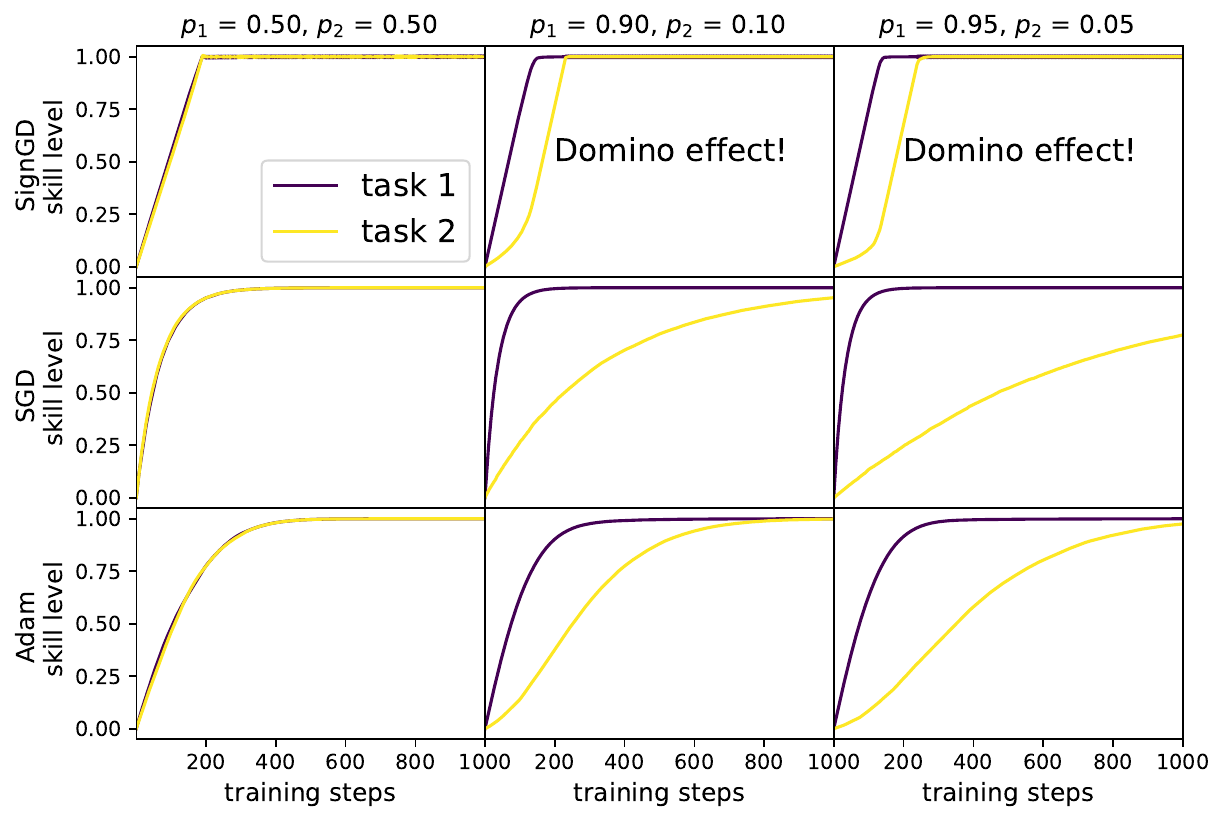}
  \caption{}
  \label{fig:2task_sub1}
\end{subfigure}%
\begin{subfigure}{0.33\textwidth}
  \centering
  \includegraphics[width=1.0\linewidth]{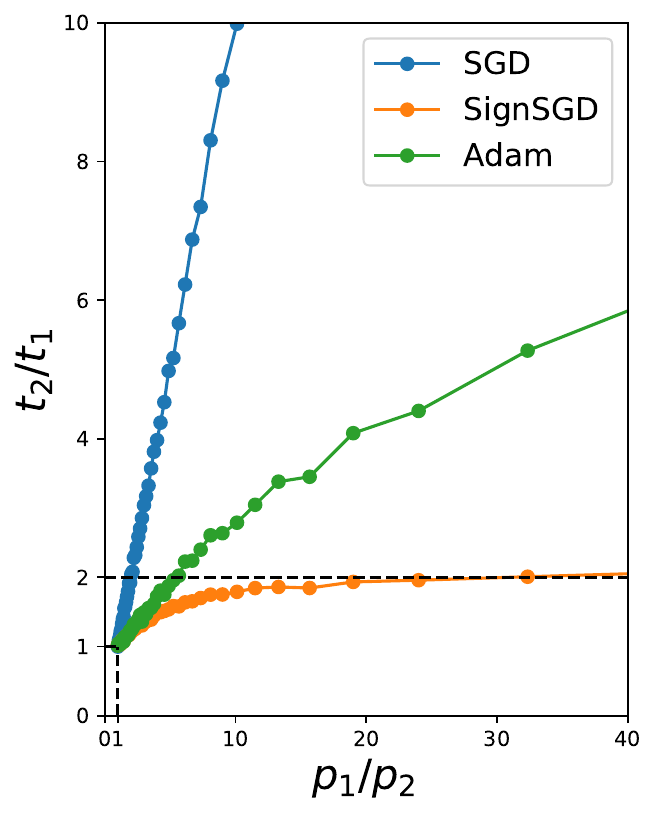}
  \caption{}
  \label{fig:2task_sub2}
\end{subfigure}
\caption{The Geometry model + SignGD leads to robust Domino effects (the second task starts to rapidly learn only after the first task completes learning), while SGD and Adam do not. (a) skill dynamics with different optimizers and different probabilities. (b) With SignGD, the learning time ratio saturates to 2 as two tasks become more unbalanced, while SGD grows linearly and Adam grows sub-linearly.}
\label{fig:2task}
\end{figure}

We now train the Geometry model and see if we can observe anything intriguing in skill dynamics (e.g., the Domino effect). Note that we have to specify optimizers to complete the Geometry model -- we choose SGD, SignGD, and Adam. For simplicity, we start with a two-task setup.

{\bf Two tasks} To gain intuition, we start with a two-task setup where task 1 and task 2 have frequencies $p_1$ and $p_2$ respectively ($p_1\geq p_2, p_1+p_2=1$). We choose the loss function to be the squared error $\mathcal{L}(s_i) = (1-s_i)^2$ to imitate regression. We choose $n_{\rm dim}=1000$, $\bm{t}_1$ and $\bm{t}_2$ are first randomly drawn from Gaussian, and then orthogonalized and normalized i.e., $||\bm{t}_1||=||\bm{t}_2||=1, \bm{t}_1\cdot\bm{t}_2=0$. We train the Geometry model with three optimizers: SGD (learning rate $3\times 10^{-2}$), SignGD ($3\times 10^{-4}$) and Adam ($3\times 10^{-4}$), and we use batch size 128. The training dynamics of both skills are shown in Figure~\ref{fig:2task} left. We test three weights: $(p_1, p_2)=(0.5,0.5), (0.9,0.1), (0.95,0.05)$. The convergence time of the second task trained with SGD is quite sensitive to $p_2$ and increases roughly $\sim 1/p_2$ (see Figure~\ref{fig:2task} right). By contrast, SignGD produces a constant delay for task 2, regardless of how small $p_2$ is. Adam's behavior lies in the middle of SGD and SignGD. Note that SignGD is a special case of Adam when $(\beta_1,\beta_2)=(0,0)$ while Adam chooses $(\beta_1,\beta_2)=(0.9,0.999)$ by default. We are particularly interested in the constant delay induced by SignGD. The second skill appears to start making rapid progress only when the first skill completes learning (Domino effect). 

\begin{figure}[t]
    \centering
    \includegraphics[width=0.8\linewidth]{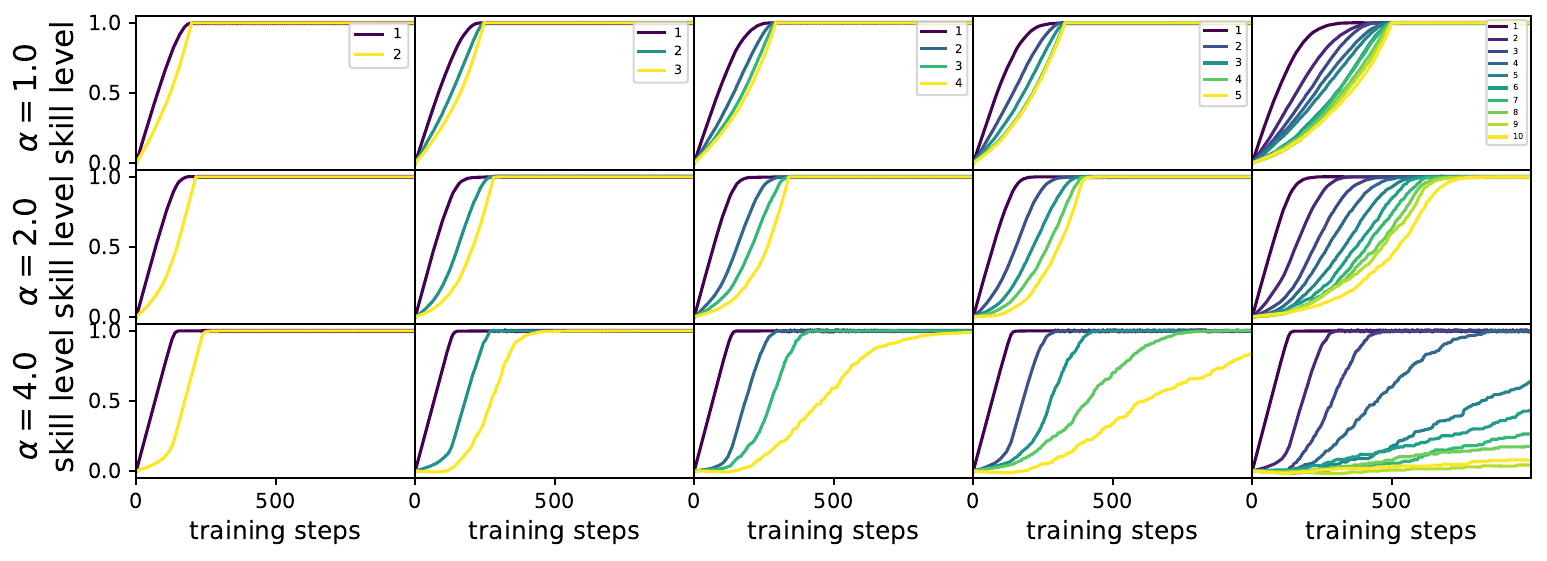}
    \includegraphics[width=0.85\linewidth]{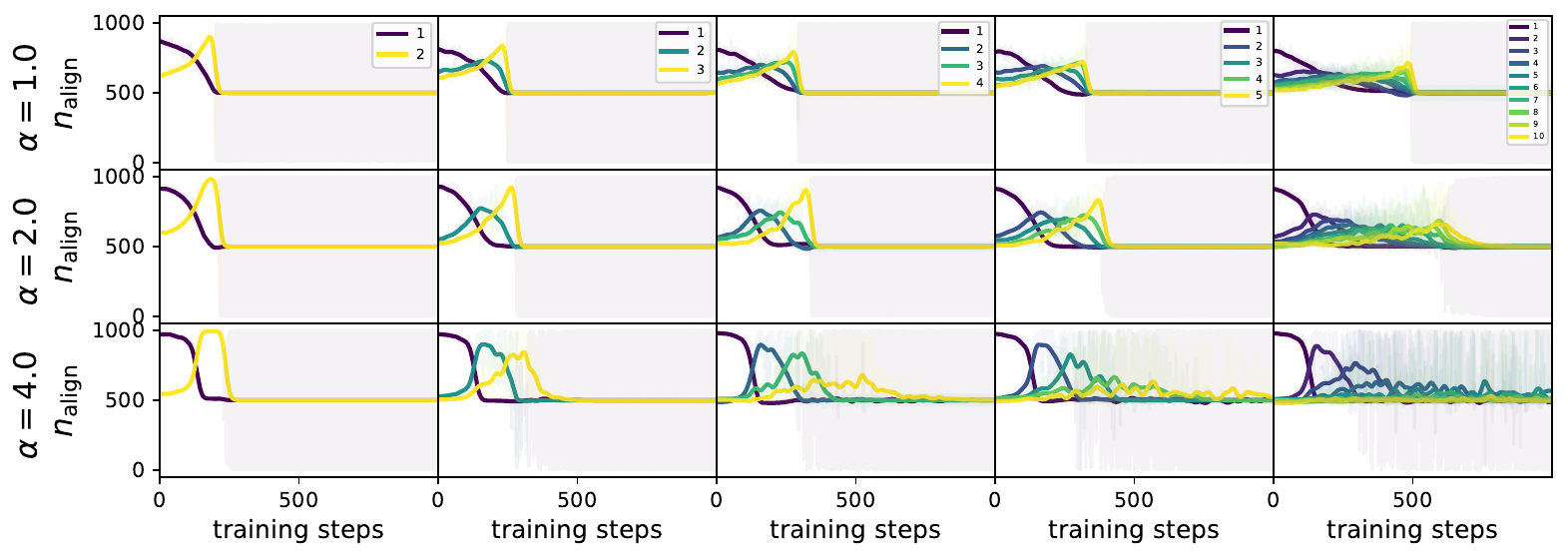}
    \caption{Results of the Geometry model. Top: Skills show sequential learning dynamics under different power law distributions $\alpha=\{1,2,4\}$ and $n_{\rm task}=\{2,3,4,5,10\}$. Bottom: the number of gradient-aligned dimensions $n_{\rm align}$ (interpreted as resources) also shows a sequential order.}
    \label{fig:nalign}
\end{figure}

{\bf A resource interpretation for the Domino effect} 
While it is true that skill vectors are orthogonal, they may still interfere through nonlinear operations in optimization. Take SignGD for example - suppose skill 1 and skill 2 have gradients $g_1$ and $g_2$ respectively, and the total gradient is $\bm{g} = \bm{g}_1 + \bm{g}_2$. 
SignGD applies element-wise sign operations to the total gradient ${\rm sign}(\bm{g})={\rm sign}(\bm{g}_1+\bm{g}_2)$. When $\bm{g}_1$ has a much larger magnitude than $\bm{g}_2$, i.e., $||\bm{g}_1||\gg ||\bm{g}_2||$, $\bm{g}_1$ will dominate $\bm{g}_2$ in almost every element and ``wins'' almost all signs, i.e., ${\rm sign}(\bm{g}_1+\bm{g}_2)\approx {\rm sign}(\bm{g}_1)$. This means SignGD will step in a direction whose elementwise signs align perfectly with those of $\bm{g}_1$ while completely ignoring $\bm{g}_2$. Only when $\bm{g}_1\approx0$ becomes almost 0 can $\bm{g}_2$ start to dominate the gradient and make progress. This is in contrast to SGD, where the update direction is $\bm{g}=\bm{g}_1+\bm{g}_2$ meaning two gradients do not interfere. To quantify this intuition, we compute the number of gradient-aligned dimensions $n_{\rm align}$ throughout training. For a gradient $\bm{g}$, its gradient with task $i$ gradient is defined as $n_{i,{\rm align}}\equiv \sum_{j=1}^{n_{\rm dim}} {\rm Sign}(\bm{g}_i)_j\cdot {\rm Sign}(\bm{g})_j$. As shown in Figure~\ref{fig:nalign} bottom (first column), we indeed see that $n_{1,{\rm align}}$ is initially high and decreases later, while $n_{2,{\rm align}}$ is initially low but increases later on. The fall of $n_{1,{\rm align}}$ coinciding with the rise of $n_{2,{\rm align}}$ has an intuitive resource interpretation: The total amount of resources is conserved (determined by $n_{\rm dim}$); skills compete for these learning resources based on their gradient magnitudes. The resource interpretation serves as the inspiration for the Resource model we proposed below in Section~\ref{subsec:effective}. 

{\bf Varying power law exponent $\alpha$ and $n_{\rm task}$.} After understanding the two-task setup, we now assume a power law distribution for many tasks, $p_i\propto i^{-\alpha}$. We present the dynamics for different $\alpha\in \{1,2,4\}$ and different $n_{\rm task}=\{2,3,4,5,10\}$ in Figure~\ref{fig:nalign}, trained with SignGD (learning rate $3\times 10^{-4}$) and batch size 128. As shown in Figure~\ref{fig:nalign} top panel, skills are indeed learned sequentially based on the order of their frequencies, corresponding to the sequence of peaks in the bottom panel. Although the resource intuition above can explain these skill dynamics qualitatively, we still do not fully understand why skill dynamics is quantitatively so. For example, the Domino effect seems to be valid only for large $\alpha$ and the first few skills, while skills in other cases display somewhat smoother learning curves. To understand these quantitative features, we propose the Resource model below.

\subsection{Introducing the Resource model}\label{subsec:effective}

{\bf Resource model} We have provided a resource interpretation to the gradient-aligned dimensions and observed that the rise of resources for one task usually corresponds to a decrease of resources for others. This suggests that the number of model parameters $n_{\rm dim}$ behaves as a resource pool that is allocated to tasks. Based on this intuition, we propose the following Resource model:

\begin{center}
\begin{tcolorbox}[colframe=black, boxrule=1pt, colback=white, width=0.95\textwidth]
{\bf Resource Model} Suppose a network is tasked with $n_{\rm task}$ tasks whose frequencies are $p_1\geq p_2\geq\cdots\geq p_{n_{\rm task}}>0$. The skill levels of these tasks are initially $s_i=0\ \forall i$, and we define unskill level $u_i\equiv 1-s_i$. The gradient magnitude of skill $i$ is defined as $N_i\equiv p_iu_i$. The resource obtained by task $i$ depends on the relative gradient magnitudes, $\tilde{N}_i\equiv\frac{N_i}{(\sum_{j=1}^{n_{\rm task}} N_j)+N_0}$, where $N_0$ characterizes the wasted resources that are not allocated to any task. The decreasing rate of unskill (or increasing rate of skill) is proportional to $\tilde{N}_i$:
\begin{equation}\label{eq:eff}
    \frac{du_i}{dt} = -\eta_{\rm eff}\tilde{N}_i=-\eta_{\rm eff}\frac{N_i}{(\sum_{j=1}^{n_{\rm task}} N_j)+N_0}=-\eta_{\rm eff}\frac{p_iu_i}{(\sum_{j=1}^{n_{\rm task}} p_ju_j)+N_0}.
\end{equation}
The effective learning rate $\eta_{\rm eff}=2\sqrt{n_{\rm dim}}\eta_{\rm geo}$, where $\eta_{\rm geo}$ is the learning rate used in the Geometry model.
\end{tcolorbox}
\end{center}
Note that the constant factor $2\sqrt{n_{\rm dim}}$ is 
obtained by synchronizing the Geometry model and the Resource model for the single-task case $N_{\rm task}=1$~\footnote{The theoretical interpretation is: the factor 2 comes from differentiating with respect to the MSE loss $\mathcal{L}(s)=(1-s)^2$. The factor $\sqrt{n_{\rm dim}}$ is because SignGD moves $\pm \eta_{\rm mech}$ along each dimension, so the total movement has length $\sqrt{n_{\rm dim}}\eta_{\rm mech}$.  }. Notice that at $t=0$, $u_1=\cdots=u_{n_{\rm task}}=1$, $\{p_i\}$ is given by data distribution. $N_0$ is the only phenomenological parameter in the Resource model -- once $N_0$ is provided, the ordinary differential equation (Eq.~\ref{eq:eff}) can be numerically solved. 

\begin{figure}[h]
    \centering
    \includegraphics[width=1.0\linewidth]{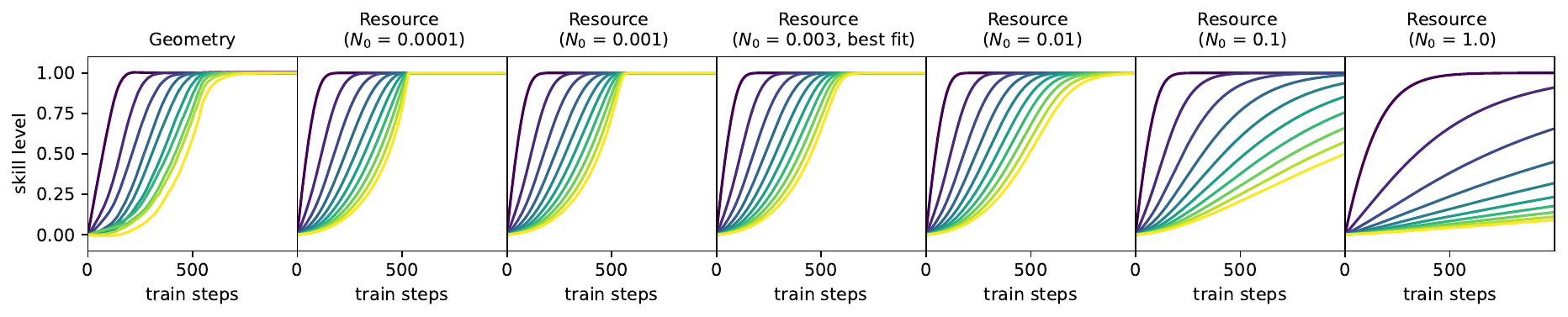}
    \caption{How $N_0$ controls the skill dynamics of the Resource model.}
    \label{fig:eff_N0}
\end{figure}

{\bf How $N_0$ controls the Resource model.} $N_0$, the only phenomenological parameter of the Resource model, can be understood intuitively as the proportion of wasted resources. As shown in Figure~\ref{fig:eff_N0}, smaller $N_0$ leads to faster learning, while larger $N_0$ delays learning. We can tune $N_0$ to make the skill dynamics induced by the Resource model match that of the Geometry model.

\begin{figure}[ht]
    \centering
    \includegraphics[width=0.8\linewidth]{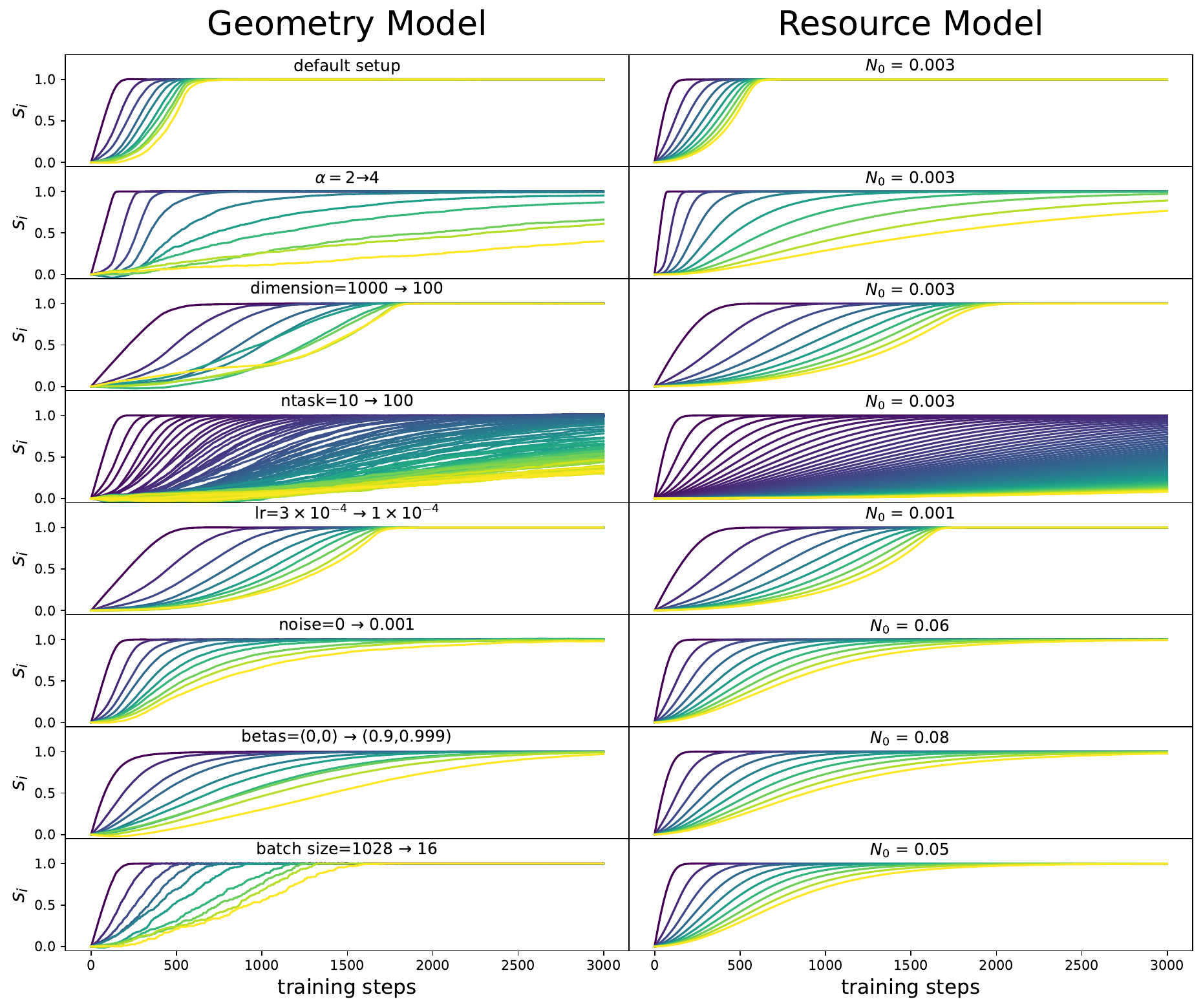}
    \includegraphics[width=0.9\linewidth]{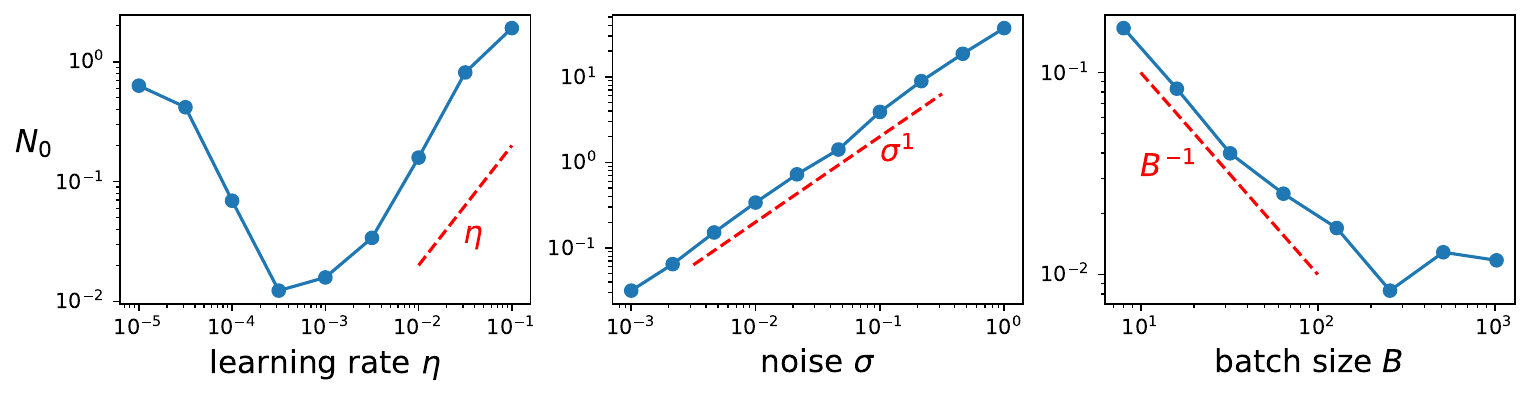}
    \caption{Overparametrized case ($n_{\rm dim}\geq n_{\rm task}$). Top: Comparing the predictions of the Geometry model (left) to the Resource model (right). Bottom: Dependence of $N_0$ on learning rate (left), noise (middle) and batch size (right).}
    \label{fig:experiment_theory_compare}
\end{figure}

{\bf How hyperparameters map to $N_0$.}
Although the Geometry model contains many low-level details (task vectors and optimizer hyperparameters), these details are all coarse-grained and are effectively represented by the phenomenological parameter $N_0$. To verify the effectiveness of the Resource model with different hyperparameters, we change optimizer parameters, finding that in all cases, there always exists an $N_0$ for each case that can induce skill dynamics similar to that of the Geometry model, as shown in Figure~\ref{fig:experiment_theory_compare}. In particular, simply changing data distribution (changing scaling law exponent from 2 to 4, changing the number of tasks from 10 to 100, changing from power law to exponential distribution) does not change the value of $N_0$, which suggests that $N_0$ does not need to be tuned every time and seems to only depend on optimization details. We find that decreasing the learning rate leads to a smaller $N_0$. This is because smaller learning rates lead to smaller bouncing back and forth (noises) for already learned skills, taking away fewer resources from skills that are being learned. Similarly, increasing noises (we manually add Gaussian noises to gradients) or reducing batch sizes lead to an increase of $N_0$. Changing the betas of the Adam optimizer could also lead to a different $N_0$. We plot how $N_0$ depends on learning rates, noises and batch sizes in Figure~\ref{fig:experiment_theory_compare} bottom.

{\bf Conserved quantities} Although it is handy to numerically solve Eq.(\ref{eq:eff}), we want to mention a nice analytical property of Eq.(\ref{eq:eff}) -- the system of $n_{\rm task}$ equations has $n_{\rm task}-1$ conserved quantities! Note that for $i\neq j$, $\frac{1}{p_iu_i}\frac{du_i}{dt}=\frac{1}{p_ju_j}\frac{du_j}{dt}$ leads to a conserved quantity $u_i^{1/p_i}-u_j^{1/p_j}=0$, or $u_1^{1/p_1}=u_2^{1/p_2}=u_3^{1/p_3}=...=u_{n_{\rm task}}^{1/p_{n_{\rm task}}}\equiv C$. This implies that learning curves should collapse when their unskill is exponentiated by a power exponent inverse to its frequency. We verify that this statement is true numerically for the Resource model in Appendix~\ref{app:collase} (Figure~\ref{fig:experiment_theory_compare_collapse} right), and approximately true for the Geometry model (Figure~\ref{fig:experiment_theory_compare_collapse} left). 

{\bf Learning time} Now $C$ is the only variable to be determined, since $u_i=C^{p_i}$. Inserting $u_i=C^{p_i}$ back to Eq.~(\ref{eq:eff}), we obtain
\begin{equation}
    \frac{dC}{dt} = -\eta_{\rm eff}\frac{C}{(\sum_{j=1}^{n_{\rm task}} p_j C^{p_j}) + N_0},
\end{equation}
which integrates to (note $C(t=0)=1$ and $C(t>0)\in (0,1)$):
\begin{equation}
    t = \underbrace{(n_{\rm task} - \sum_{j=1}^{n_{\rm task}}C^{p_j})/\eta_{\rm eff}}_{t_{\rm task}} + \underbrace{(- N_0{\rm ln}C)/\eta_{\rm eff}}_{t_{\rm waste}}.
\end{equation}
This means that: to decrease $C$ from 1 to $C<1$, it takes time $t$ which consists of two parts: $t_{\rm task}$ and $t_{\rm waste}$. Note that $\sum_{j=1}^{n_{\rm task}} C^{p_j}\approx A(\alpha, C)n_{\rm task}$ is proportional to $n_{\rm task}$ while the factor $A(\alpha,C)\in (0,1)$ depends on $C$ and $\alpha$ (power exponent of $p_j\propto j^{-\alpha}$). As a result, $t_{\rm task}=(1-A(\alpha, C))n_{\rm task}\propto n_{\rm task}$, i.e., linear to the number of tasks. $t_{\rm waste}=(-N_0{\rm ln}C)/\eta_{\rm eff}\propto N_0$, i.e., linear to wasted resources. 

{\bf Remark on learning rate decay} $N_0$ is an increasing function of learning rate $\eta$ when $\eta$ is not too small (Figure~\ref{fig:eff_N0} bottom left). Considering this dependence $N_0=N_0(\eta)$, the right-hand side of Eq.~(\ref{eq:eff}) is not necessarily monotonic to $\eta$. Phenomenologically when $N_0 \propto \eta^2$, the optimal learning rate (that maximizes the magnitude of the right-hand side of Eq.~(\ref{eq:eff}) is $\eta^*\propto \sqrt{\sum_{j=1}^{n_{\rm task}}p_ju_j}$, which is decreasing in time. This explains the learning rate decay strategy commonly adopted in machine learning. When $N_0 \propto \eta$, the decrease rate monotonically increases (and finally saturates) with $\eta$, where learning rate decay is unnecessary in the continuous limit described by Eq.~(\ref{eq:eff})~\footnote{In practice, the optimization dynamics is gradient descent rather than gradient flow - such discreteness gives another reason for reducing learning rates towards the end of learning.}. This suggests that in large language models, it is probably the case that $N_0(\eta)$ grows faster than linear, to justify the need for learning rate decay. 

\subsection{The Underparametrized regime: $n_{\rm task}> n_{\rm dim}$}\label{subsec:underparam}

\begin{figure}[ht]
    \centering
    \includegraphics[width=0.9\linewidth]{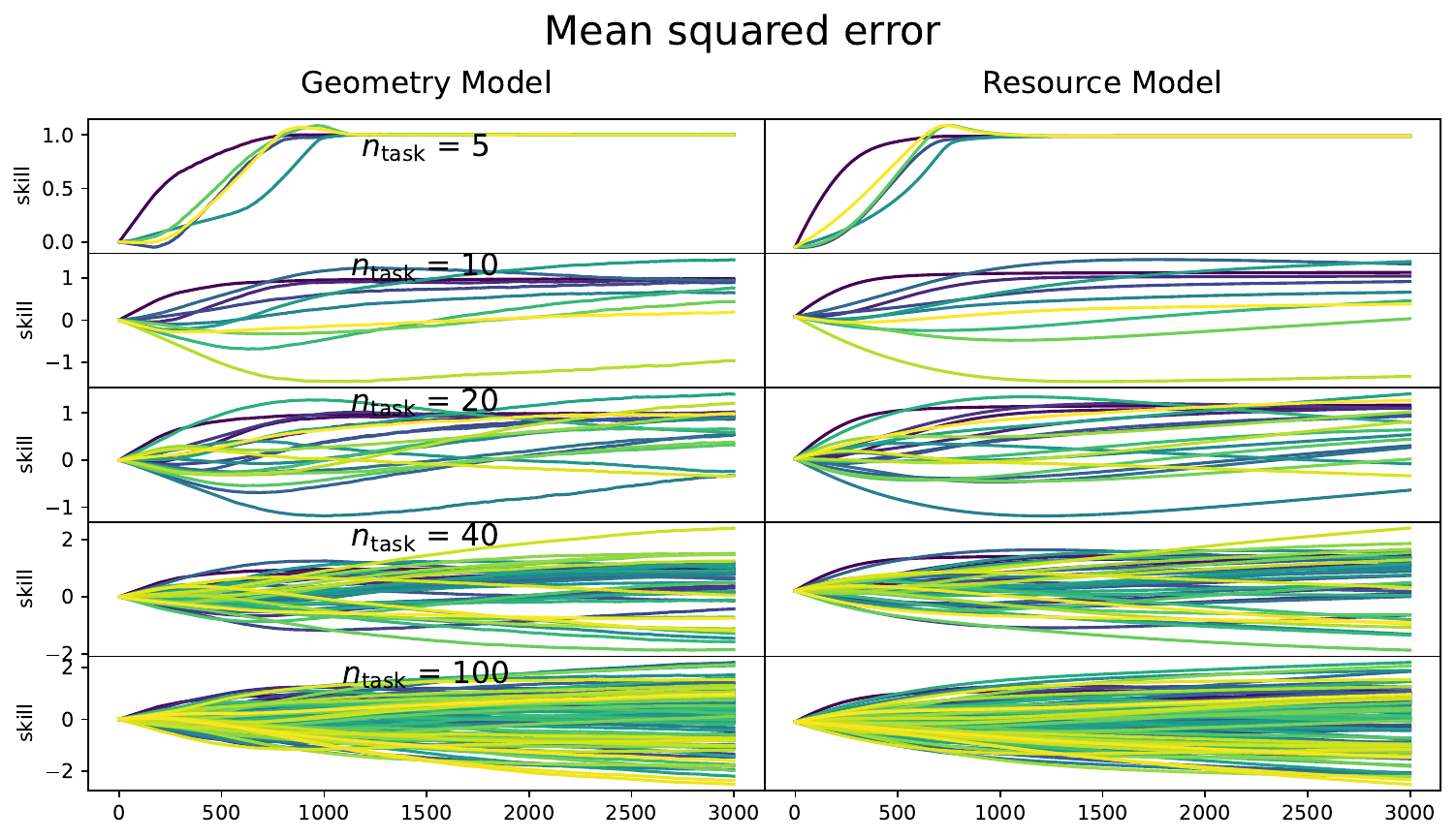}
    \includegraphics[width=0.9\linewidth]{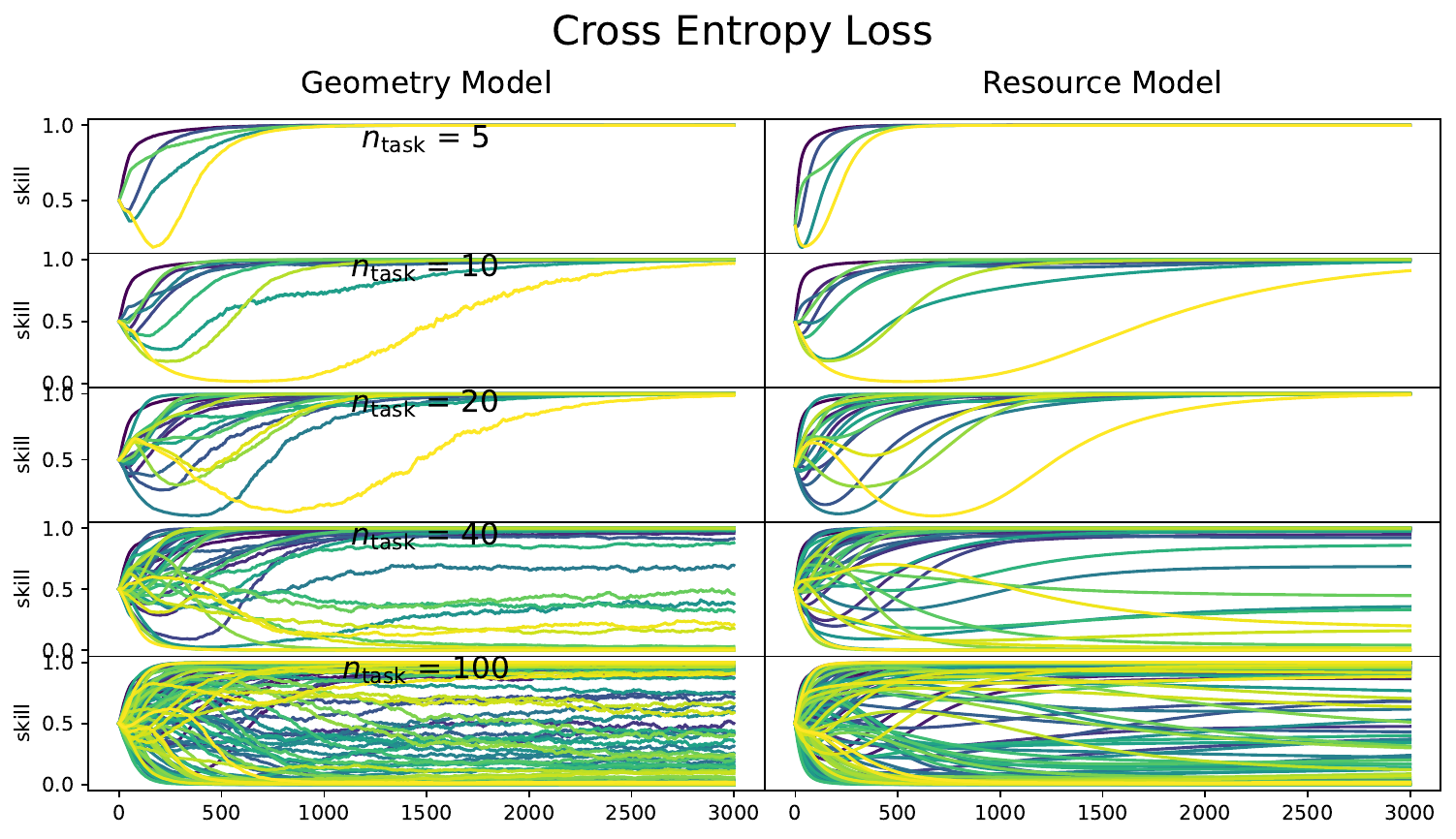}
    \caption{Underparametrized case ($n_{\rm dim} < n_{\rm task}$): Comparing the Geometry model to the Resource model. Top: regression (mean squared error). Bottom: classification (cross-entropy loss).}
    \label{fig:underparam}
\end{figure}

The philosophy behind the ``scaling is all you need'' narrative lies in the assumption that the current large language model, albeit having billions of parameters, still falls short of capability in the face of complex language data. To put it another way, we are still in the regime of under-parametrization for language modeling. 
In the under-parameterized regime, not all skills can be learned, and interference of skills becomes inevitable because it is impossible to get $n_{\rm task}$ orthogonalized vectors in an $n_{\rm dim} < n_{\rm task}$ dimensional space. 

\subsubsection{The Geometry model}
To study the under-parametrized regime of the Geometry model, no extra treatment is needed except for setting $n_{\rm task}>n_{\rm dim}$, and no longer orthogonalization of task vectors. Each task vector $\bm{t}_i$ is a vector of unit length with a random angular direction. The inner product between two task vectors indicates whether they are collaborative (when the inner product is positive) or competitive (when the inner product is negative). We analyze both the  regression setup (using squared loss $\mathcal{L}(s)=(1-s)^2$) and the classification setup (using cross-entropy loss $\mathcal{L}(s)=-{\rm log}(\sigma(s))$ with $\sigma(s)=\frac{1}{1+e^{-s}}$). We set $n_{\rm dim}=10$, and vary $n_{\rm task}=\{5,10,20,40,100\}$. We use the SignGD optimizer (learning rate $3\times 10^{-4}$) with a batch size of 128. Skill dynamics are shown in Figure~\ref{fig:underparam} top left (Mean squared error) and bottom left (Cross entropy loss). It is expected that as the number of tasks increases, some tasks cannot be learned at all, or even become worse than random guessing due to the negative correlation with strong skills that dominate the learning dynamics. The correlations bring in non-monotonic dynamics, which are new rich dynamics that were not observed in the overparametrized regime.

\subsubsection{The Resource model}

To characterize the under-parametrized regime, extra treatment is needed for the Resource model -- how we could model correlations between different skills. It turns out we could use the inner products from the Geometry model. We define the correlation matrix $\bm{C}$ with $\bm{C}_{ij}=\bm{t}_i\cdot \bm{t}_j$. Note that $\bm{C}_{ij}$ are not new phenomenological parameters since they can be determined by task vectors of the Geometry model. It is clear that $\bm{C}_{ii}=1$. The resource model in the overparametrized regime (section~\ref{subsec:effective}) was a special case where all non-diagonal terms are zero, $\bm{C}_{ij}=0, \forall i\neq j$. Now we present more general Resource models that capture the correlations between tasks.

{\bf Regression}  We first present the Resource model with correlation in the regression setup. Correlations from other skills enter into the nominator of the right-hand side:
\begin{equation}\label{eq:eff_underparam_mse}
    \frac{du_i}{dt} =-\eta_{\rm eff}\frac{\sum_{k=1}^{n_{\rm task}}C_{ik}p_ku_k}{(\sum_{j=1}^{n_{\rm task}} p_ju_j)+N_0}.
\end{equation}
It is easy to check that when $C$ is an identity matrix, Eq.~(\ref{eq:eff_underparam_mse}) degrades to Eq.~(\ref{eq:eff}). The skill dynamics is shown in Figure~\ref{fig:underparam} top right, manifesting good agreement with the Geometry model. 

{\bf Classification} Since language modeling is formulated as next-token prediction, which is a classification problem, it is more realistic to use the cross entropy loss rather than the mean squared error in the Resource model. Note that the term $u_k\equiv 1-s_k$ on the right-hand side of Eq.~(\ref{eq:eff_underparam_mse}) is simply the gradient of the squared error $\mathcal{L}(s_k)=(1-s_k)^2$. By this analogy, we can replace $u_k$ with the gradient of the cross-entropy loss $\mathcal{L}(s_k)=-{\rm log}(\sigma(s_k))$, which is $u_k=\frac{d\mathcal{L}(s_k)}{ds_k}=\frac{e^{-s_k}}{1+e^{-s_k}}$, giving
\begin{equation}\label{eq:eff_underparam_ce}
    \frac{ds_i}{dt} =\eta_{\rm eff}\frac{\sum_{k=1}^{n_{\rm task}}C_{ik}p_k\frac{e^{-s_k}}{1+e^{-s_k}}}{(\sum_{j=1}^{n_{\rm task}} p_j\frac{e^{-s_j}}{1+e^{-s_j}})+N_0}.
\end{equation}
The skill dynamics are shown in Figure~\ref{fig:underparam} bottom right, also manifesting good agreement with the Geometry model. In particular, the non-monotonic skill learning dynamics are also captured by the Resource model. The analytical behavior of the Resource model in the underparametrized regime is left for future work.

\subsection{Introducing the Domino Model}\label{subsec:domino}

\begin{figure}
    \centering
    \includegraphics[width=0.9\linewidth]{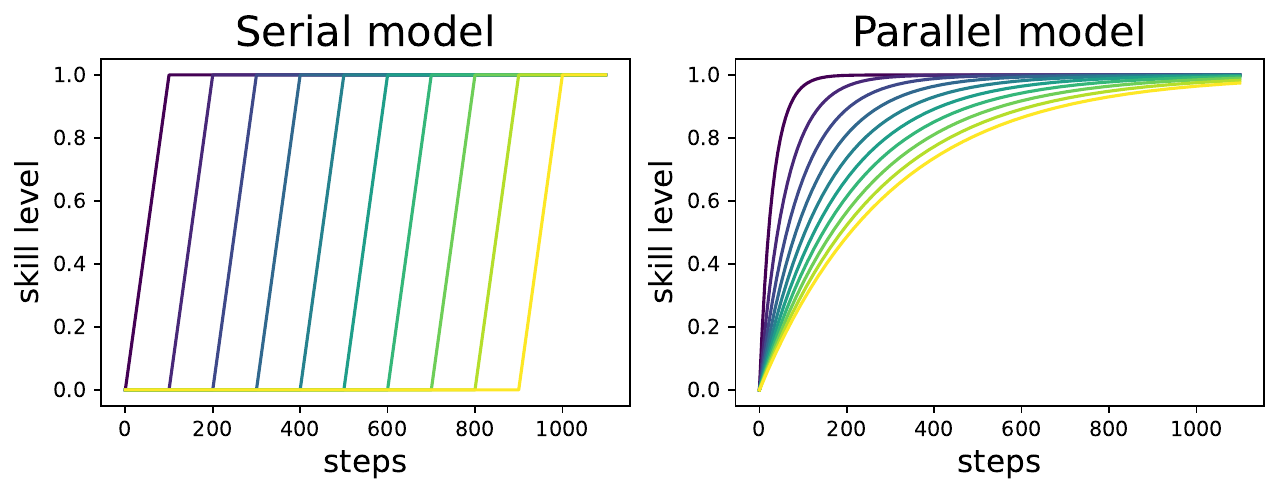}
    \caption{Left: Serial model (Domino model) - skills are learned in a strict sequential order. The time to learn the $n^{\rm th}$ task is $t_n^{\rm D}\propto n$.  Right: Parallel model (Quanta model + $t\sim p^{-1}$ assumption) - skills are learned independently. The time to learn the $n^{\rm th}$ task is $t_n^{\rm Q}\propto p_n^{-1}\propto n^{\alpha}$.}
    \label{fig:domino}
\end{figure}

We first introduce the Domino model without justification:

\begin{center}
\begin{tcolorbox}[colframe=black, boxrule=1pt, colback=white, width=0.95\textwidth]
{\bf Domino Model} Suppose a network is tasked with $n_{\rm task}$ tasks whose frequencies are $p_1\gg p_2\gg p_3 \gg \cdots \gg p_{n_{\rm task}}>0$. The skill levels of these tasks are initially $s_i=0\ \forall i$. The skills are learned in a strict sequential order. Suppose it takes $t_0$ time to learn one task, then skill $n$ has level (illustrated Figure~\ref{fig:domino} left):
\begin{equation}
    s_n=
    \begin{cases}
      0, & \text{if}\ t \leq nt_0, \\
      1, & \text{if}\ t > (n+1)t_0, \\
      t/t_0-n, & \text{if}\ nt_0<t\leq (n+1)t_0, \\
    \end{cases}
  \end{equation}
\end{tcolorbox}
\end{center}

We now show that the Domino model can be derived from the Resource model as a special case. Although the Resource model is already quite simple, solving the differential equations Eq.~(\ref{eq:eff}) is still cumbersome and may prohibit intuitive thinking. We want to further impose a few simplifying assumptions so that Eq.~(\ref{eq:eff}) can simplify to something that can be solved without any effort. We make two assumptions: (1) {\bf No waste assumption}, i.e., $N_0=0$; (2) {\bf Strong hierarchy assumption}, i.e., $p_1\gg p_2\gg p_3\gg\cdots$. With these two assumptions, initially, we have 
\begin{equation}
    \frac{du_1}{dt} = -\eta_{\rm eff}, \quad \frac{du_i}{dt} = 0\ (i>1),
\end{equation}
which means the first skill is learned with a constant speed. It takes $t_0 = 1/\eta_{\rm eff}$ to reduce $u_1$ from 1 to 0. When $u_1$ becomes 0, it is time for the second task to dominate the dynamics, i.e.,
\begin{equation}
    u_1=0, \quad \frac{du_2}{dt} = -\eta_{\rm eff}, \quad \frac{du_i}{dt} = 0\ (i>2),
\end{equation}
which means the second skill is learned with a constant speed, again taking time $t_0=1/\eta_{\rm eff}$ to complete learning before the third skill kicks off. This chain of reaction goes on until all the skills are learned (see Figure~\ref{fig:domino} for an illustration). This model is called the Domino model because it obeys the exact order of learning, the same as the sequential falls of Domino cards. In practice, the model may only be capable of learning $n'_{\rm task}<n_{\rm task}$ tasks. In this case, the Domino chain goes on until task $n'_{\rm task}$ and the rest tasks remain unlearned forever.

{\bf Learning time} To learn $n_{\rm task}$ skills, it takes time $t = n_{\rm task}t_0\propto n_{\rm task}$, i.e., linear to the number of tasks, which agrees with the Resource model. Although this result sounds intuitive, is this the best scaling we can hope for? In Section~\ref{sec:modularity}, we show how to reduce learning time from $O(n_{\rm task})$ to $O(\sqrt{n_{\rm task}})$ by leveraging modularity.

\section{Implications for Neural Scaling Laws}\label{sec:nsl}

Neural scaling laws, the phenomenon that model performance progressively improves as resources (data, parameters, compute) scale up, are the main driving force for today's deep learning~\cite{kaplan2020scaling,hoffmann2022training}. It is thus intriguing to understand the mechanisms behind the neural scaling laws. In particular, as formulated by Michaud et al.~\cite{michaud2024quantization}, we are interested in how $\alpha_N$ ($\ell\sim n_{\rm dim}^{-\alpha_N}$, exponent against the number of parameters) and $\alpha_S$ ($\ell\sim S^{-\alpha_S}$, exponent against the number of steps or data) are dependent on $\alpha$ ($\alpha$ specifies the data distribution $p_j\propto j^{-\alpha}$). We first revisit the Quanta model in~\cite{michaud2024quantization}, and then discuss how our models can enrich the Quanta model family. 

\subsection{Revisit the Quanta model and compare with the Domino model}

The premise of the Quanta model~\cite{michaud2024quantization} is the hypothesis that ``network knowledge and skills are quantized
into discrete chunks (quanta).'', which is also shared by our models. The quanta model can provide accurate predictions for $\alpha_N$ but usually underestimates $\alpha_S$. 

{\bf Predicting $\alpha_N$}. The Quanta model assumes that each task requires a fixed amount of resources ($C$ parameters) hence a network with $n_{\rm dim}$ parameters is able to learn $n_0\equiv n_{\rm dim}/C$ tasks. The unlearned tasks contribute to the loss function (taking $n_{\rm task}\to\infty$): $\ell =  \sum_{i=n_0}^{\infty} p_i \sim \sum_{i=n_0}^{\infty} i^{-\alpha}\sim n_0^{-\alpha+1} \sim n_{\rm dim}^{-(\alpha-1)} \ (\alpha>1)$, giving the parameter scaling exponent $\alpha_N^Q = \alpha - 1$~\footnote{Note that our $\alpha$ is equivalent to $\alpha+1$ in~\cite{michaud2024quantization}. In particular, we assume data distribution $p_n\propto n^{-\alpha}$ while~\cite{michaud2024quantization} assumes $p_n\propto n^{-(\alpha+1)}$.}. The Domino model has the same story (in fact $C=1$, since each direction corresponds to a skill), also predicting $\alpha^D_N = \alpha-1$.


{\bf Predicting $\alpha_S$}. The Quanta model~\cite{michaud2024quantization} additionally assumes that: the time to learn task $n$ is $t^Q_n \propto 1/p_n\propto n^{\alpha}$. This means that at step $S\propto n^{-\alpha}$, $n$ tasks are learned, leading to loss $\ell \propto n^{-(\alpha-1)} \propto S^{-\frac{\alpha-1}{\alpha}}$, giving $\alpha_S^D = \frac{\alpha-1}{\alpha}$. However, the relation $t_n\propto p_n^{-1}$ only makes sense when SGD is used as the optimizer: the gradient signal is proportional to frequency, and learning time is inversely proportional to the gradient signal. However, when adaptive optimizers like SignGD are used, 
the Domino model suggests that $t^D_n\propto n$. This means that at step $S \propto n$, $n$ tasks are learned, leading to loss $\ell \propto n^{-(\alpha-1)} \propto S^{-(\alpha-1)}$, giving $\alpha^Q_S = \alpha-1$. 

{\bf Remarks} For $\alpha > 1$, the Quanta model predicts that $\alpha_N > \alpha_S$, while the Domino model predicts $\alpha_N = \alpha_S$. Interestingly, this seems more in line with published scaling laws results. For instance, while the original DeepMind ``Chinchilla'' paper published fits of $\alpha_N \approx 0.34$ and $\alpha_S \approx 0.28$, a recent replication~\cite{besiroglu2024chinchilla} found a better fit to the original data with $\alpha_N \approx 0.34$ and $\alpha_S \approx 0.37$. To be clear, we are not claiming that the Domino model fully describes scaling behavior. However, if one aims to build a more realistic model and wants to incorporate the speedup effect of adaptive optimizers, the Domino model can be used for this purpose. In fact, the Domino model (like the Quanta model) gives $\alpha_N=0$ for $\alpha=1$, clearly disagreeing with observations. Below we show that the Geometry model can produce more realistic scaling exponents for $\alpha = 1$.

\subsection{The Geometry Model}

\begin{figure}[ht]
\centering
\begin{subfigure}{.55\textwidth}
  \centering
  \includegraphics[width=1.0\linewidth]{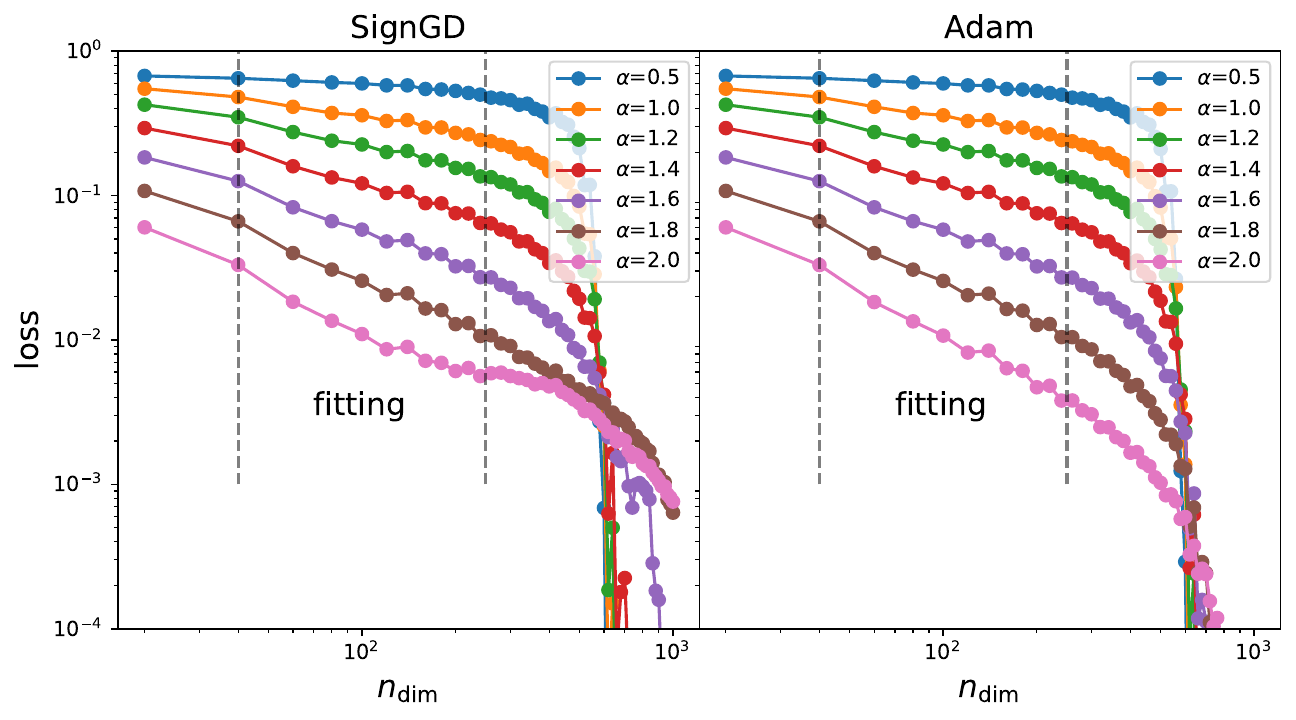}
  \caption{}
  \label{fig:nsl_sub1}
\end{subfigure}%
\begin{subfigure}{0.38\textwidth}
  \centering
  \includegraphics[width=1.0\linewidth]{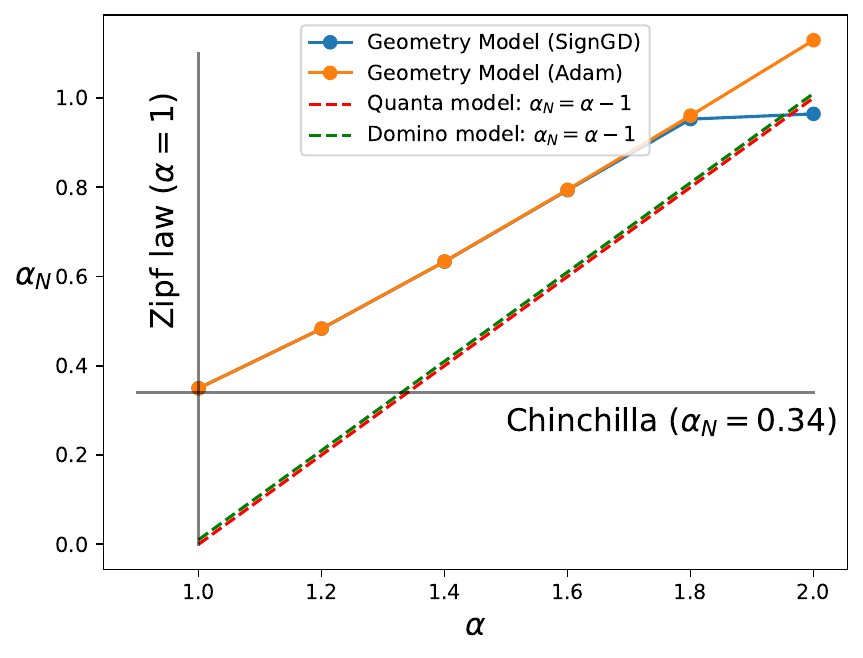}
  \caption{}
  \label{fig:nsl_sub2}
\end{subfigure}
\begin{subfigure}{.55\textwidth}
  \centering
  \includegraphics[width=1.0\linewidth]{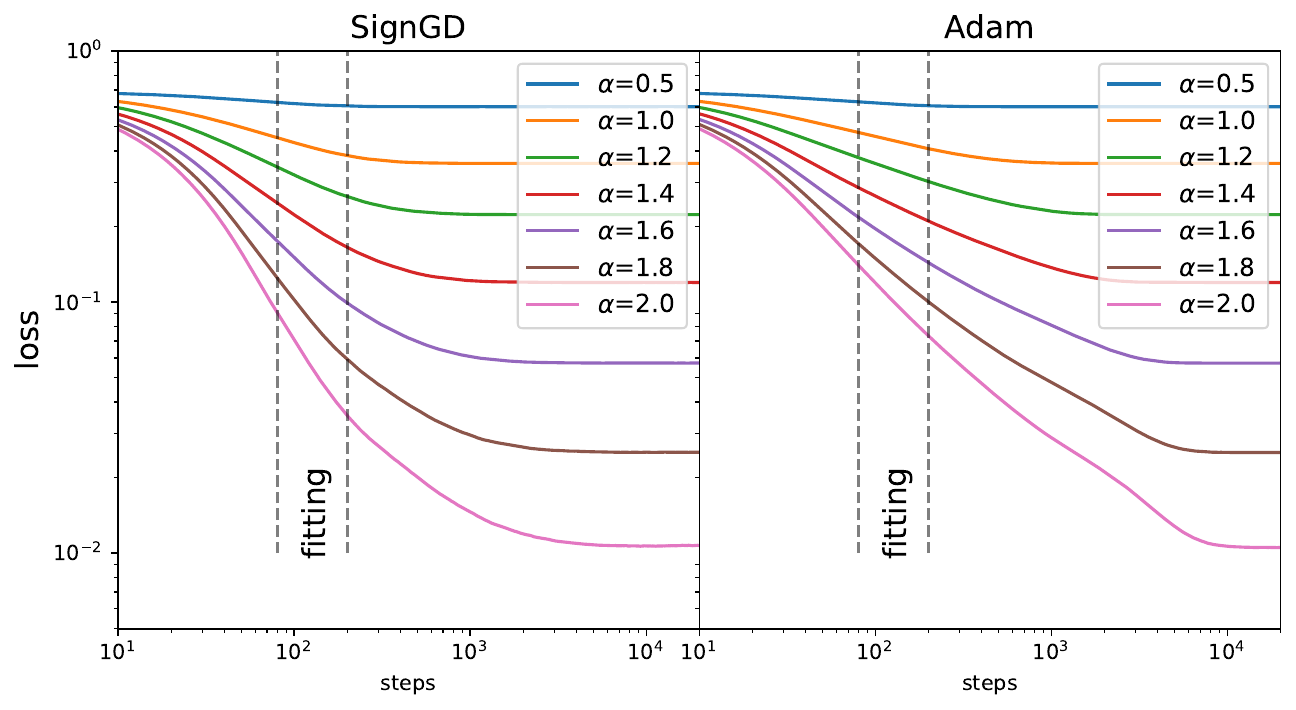}
  \caption{}
  \label{fig:nsl_sub3}
\end{subfigure}%
\begin{subfigure}{0.38\textwidth}
  \centering
  \includegraphics[width=1.0\linewidth]{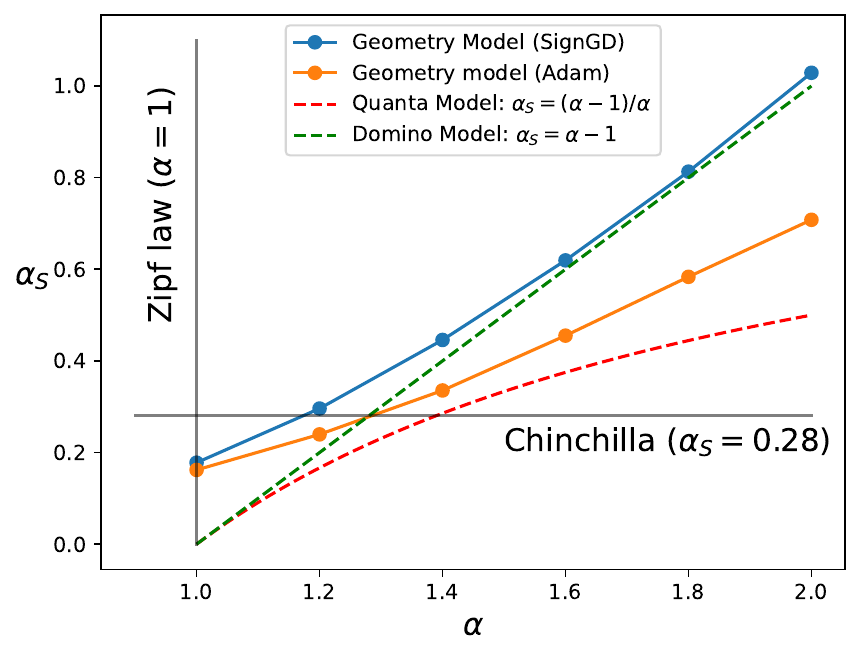}
  \caption{}
  \label{fig:nsl_sub4}
\end{subfigure}
\caption{Neural scaling laws emerge in the Geometry model. (a) loss against dimensions $\ell(n_{\rm dim})$; (b) the parameter scale exponent $\alpha_N$ against $\alpha$. (c) loss against steps $\ell(S)$; (b) the parameter scale exponent $\alpha_S$ against $\alpha$. In (b) and (d), the Geometry model gives good predictions about Chinchilla scaling exponents, assuming that data is Zipfian distribution $(\alpha=1)$.}
\label{fig:nsl}
\end{figure}

For simplicity, we use the setup in Section~\ref{subsec:underparam}, where we assume skills are independent, but since the model is underparametrized, task vectors could have positive or negative correlations with each other. We set $n_{\rm task}=1000$
, varying $n_{\rm dim}$ and data exponent $\alpha$. We train Geometry models with SignGD or Adam with a learning rate of 0.01 for 100000 steps. 

{\bf Predicting $\alpha_N$.} The loss is a power law function of $n_{\rm dim}$: $\ell = A{n_{\rm dim}^{-\alpha_N}}$. The $\ell(n_{\rm dim})$ relation is shown in Figure~\ref{fig:nsl} (a), and the extracted scaling exponent $\alpha_N(\alpha)$ is shown in (b). For reference, we also plot out $\alpha_N = \alpha-1$, which is what the Quanta model (and the Domino model) would predict. The Geometry model systematically gives a larger $\alpha_N$ than the Quanta model. In particular, for $\alpha=1$, the Geometry model gives $\alpha_N\approx 0.34$, which happens to be the scaling exponent observed in the Chinchilla models~\cite{hoffmann2022training}. By contrast, the Quanta and the Domino models would predict $\alpha_N=0$, which is far off. 

Besides the power law scaling behavior in the early stage $n_{\rm dim}<250$, There exists a critical point around $n_{\rm dim}\sim 600$ when the loss suddenly decreases. The existence of such a critical point is expected since when $n_{\rm dim} \geq n_{\rm task}=1000$, the model can learn all skills, resulting in zero loss. 
The left shift of the critical point might be due to superposition. We do not expect that the critical phenomenon (if truly meaningful) is observable in the current scales of large language models, since we conjecture that today's language models are still in the underparametrized regime (i.e., cannot overfit to all language data).    

{\bf Predicting $\alpha_S$.} The loss is a power law of training steps $S$: $\ell=BS^{-\alpha_S}$. The $\ell(S)$ relation is shown in Figure~\ref{fig:nsl} (c), and the extracted scaling exponent $\alpha_S(\alpha)$ is shown in (d). For reference, we also plot out $\alpha_N^Q = (\alpha-1)/\alpha$ (Quanta model) and $\alpha_N^D=\alpha-1$ (Domino model). There are a few interesting observations: (1) The Geometry model + SignGD results in the $\alpha_S(\alpha)$ dependence agreeing with the Domino model (especially for large $\alpha$, when tasks have a stronger hierarchy, an assumption of the Domino model). (2) Adam results in smaller $\alpha_S$ than SignGD. This suggests that optimization details can affect $\alpha_S$ to a great extent, which justifies the recent exploration of new optimizers for speeding up the training of large models. (3) The Quanta model systematically underestimates $\alpha_S$ (also observed by the Quanta paper~\cite{michaud2024quantization}).


\subsection{Multitask sparse parity}

\begin{figure}[ht]
    \centering
    \includegraphics{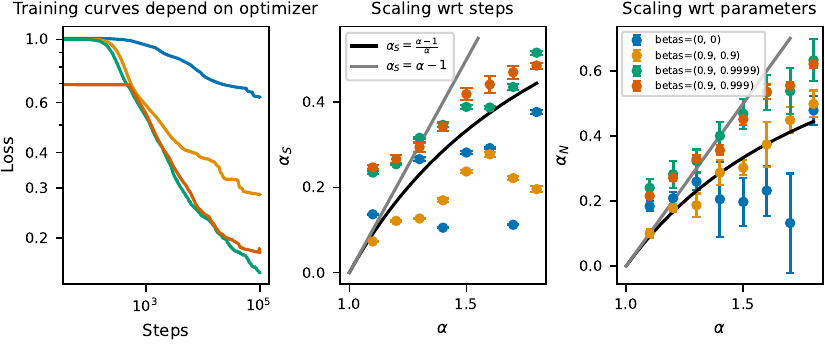}
    \caption{We study neural scaling laws on multitask sparse parity with varying choices of Adam optimizer $\beta_1, \beta_2$. We show training curves (\textbf{Left}) for the different optimizer hyperparameters for a network with 300 hidden neurons. Measuring the slope of the scaling curve wrt steps (between $10^3$ and $10^4$ steps), we show how $\alpha_S$ varies with $\alpha$ (\textbf{Center}). We also study how scaling wrt parameters (holding steps constant at $10^5$) depends on optimizer, and show how $\alpha_N$ varies with $\alpha$ (\textbf{Right}). Error bars were computed using a jackknife procedure, excluding one checkpoint or network size at a time to estimate the uncertainty in the power law fit parameters.}
    \label{fig:multitask-sparse-parity-step-parameter-scaling-tripanel}
\end{figure}

We also perform experiments training real neural networks on the multitask sparse parity task introduced by~\cite{michaud2024quantization}. 

{\bf Problem setup} Multitask sparse parity is a binary classification problem on binary strings. Input bit strings are $n_\text{tasks} + n$ bits long, where there are $n_\text{tasks}$ subtasks (skills) and $n$ is another parameter that regulates the complexity of the subtasks. For each $i \in \{1, \ldots, n_\text{tasks}\}$, we choose a random subset of $k$ of the trailing $n$ bit indices $S_i$. Only one bit in the first $n_\text{tasks}$ bits is $1$ on any input, and the rest are $0$. When bit $i$ is $1$, the label for that bit string is the parity (sum modulo 2) of the bits $S_i$. The trailing $n$ bits are sampled uniformly. For each skill $i$, the network must learn from examples to compute the parity from the correct subset of the trailing $n$ bits. We impose a power law distribution over the frequencies that the various skills occur: the probability that bit $i$ (among the first $n_\text{tasks}$ bits) is $1$ is $p_i \propto i^{-\alpha}$. We train ReLU MLPs with a single hidden layer with $n_\text{tasks} = 500$, $n=100$, $k=3$, with a batch size of 20000.

{\bf Results} We train with various configurations of the Adam optimizer $\beta_1, \beta_2$ parameters. We train with $(\beta_1, \beta_2) = (0, 0), (0.9, 0.9), (0.9, 0.999)$ and $(0.9, 0.9999)$. Figure~\ref{fig:multitask-sparse-parity-step-parameter-scaling-tripanel} shows scaling results in steps and also in the number of network parameters. Overall, we notice that scaling exponents depend to some extent on the choice of optimizer, a detail which was not considered in the original analysis of~\cite{michaud2024quantization}. For instance, with betas of $(0.9, 0.9)$, the scaling law wrt parameters seems to have a consistently lower exponent $\alpha_N$ than with the default Adam betas $(0.9, 0.999)$ or our experiments with betas $(0.9, 0.9999)$. This result emphasizes the importance of optimization considerations in scaling laws analysis. Note however that for multitask sparse parity, for no choice of Adam parameters do we find that the relationship $\alpha_S \approx \alpha$ and $\alpha_N \approx \alpha$ holds precisely. Instead, for the most performant choices of Adam parameters, we find that $\alpha_S$ and $\alpha_N$ tend to be between the lines $\frac{\alpha - 1}{\alpha}$ predicted by the quanta model and $\alpha - 1$ predicted by the Domino model.


\section{Implications for Optimization}\label{sec:optimization}

We have found that different optimizers (or the same optimizer with different hyperparameters) can lead to quite different training dynamics in the Geometry model. It is thus interesting to ask: can these observations or insights be generalized to real-world training? If this is the case, we can achieve faster idea iteration -- instead of training big models for a few days, one can experiment with simpler models in a few seconds or minutes and get transferable insights. We investigate how these insights can apply grokking (Section~\ref{subsec:grokking}), data weighting (Section~\ref{subsec:weighting}) and optimizers (understanding recently proposed optimizers in Section~\ref{subsec:optimizer}). But first, we present a simple quadratic loss function that can demonstrate the Domino effect in Section~\ref{subsec:quadratic}.

\subsection{Quadratic losses}\label{subsec:quadratic}

\begin{figure}
    \centering
    \includegraphics[width=1.0\linewidth]{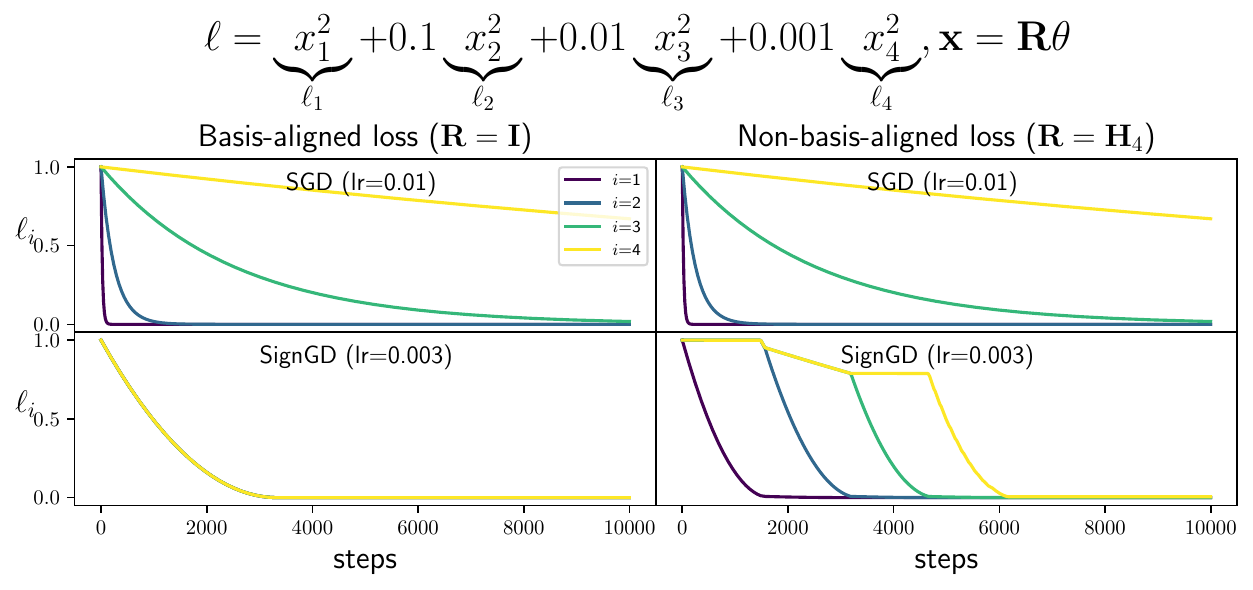}
    \caption{A quadratic loss with a strong hierarchy. We study two cases: when the loss is basis-aligned (eigendirections aligned with parameters) or non-basis-aligned. $\bm{H}_4$ standards for the $4\times 4$ Hadamard matrix. SGD induces the same dynamics for both cases. SignGD synchronizes all losses in the basis-aligned case, while displaying the Domino effect in the non-basis-aligned case.}
    \label{fig:quadratic-loss-optimization}
\end{figure}

In Section~\ref{subsec:overparam}, we have shown that the Geometry model is able to induce the Domino effect. The Geometry model is still a bit complicated though; to further simplify, we present a simple quadratic loss function that can manifest the Domino effect. The loss function is quadratic with strong hierarchies:
\begin{equation}
    \ell = \underbrace{x_1^2}_{\ell_1} + 0.1\underbrace{x_2^2}_{\ell_2} + 0.01\underbrace{x_3^2}_{\ell_3} + 0.001\underbrace{x_4^2}_{\ell_4}.
\end{equation}
Note that $\bm{x}=(x_1,x_2,x_3,x_4)^T$ are not necessarily the parameters to be optimized by optimizers. We assume $\bm{\theta}$ are the parameters that are directly optimized by optimizers, and $\bm{x}=\bm{R}\bm{\theta}$, where $\bm{R}$ is a rotation matrix. When $\bm{R}=\bm{I}$ (identity matrix), we say the loss is basis-aligned, because the eigen-directions correspond to the optimized parameters. We choose the initial point to be $\bm{x}_0=(1,1,1,1)^T$. Otherwise if $\bm{R}\neq \bm{I}$, we call the loss non-basis-aligned. In particular, we choose $\bm{R}=\bm{H}_4$ (the $4\times 4$ Hadamard matrix $\bm{H}_4=((1,1,1,1),(1,1,-1,-1),(1,-1,1,-1),(1,-1,-1,1))$). We choose the initial point to be $\bm{x}_0'=(1,1,1,-1)^T$ (since $\bm{H}\bm{x}_0'=\bm{x}_0$). SGD is invariant to rotations, as shown in Figure~\ref{fig:quadratic-loss-optimization} top two panels. However, SignGD is sensitive to rotations (bottom two panels), due to element-wise operations generally in adaptive optimizers. When the loss is basis-aligned, four sub-losses decrease at the same pace due to elementwise normalization. By contrast, when the loss is non-basis-aligned, four sub-losses display a Domino effect -- one loss starts to decrease only when the previous loss decreases to around zero.

\subsection{Grokking}\label{subsec:grokking}

Grokking refers to the phenomenon when generalization happens long after memorization. The learning dynamics is featured by a fast overfitting phase and a slow phase transiting to generalization. Most investigations on grokking use the Adam optimizer, because of its popularity in machine learning. However, we observed in Figure~\ref{fig:2task} that SignGD can outperform Adam for the specific case, which inspires us to try SignGD in the grokking setup, and see how it compares with Adam.

\begin{figure}[ht]
\centering
\begin{subfigure}{.49\textwidth}
  \centering
  \includegraphics[width=1.0\linewidth]{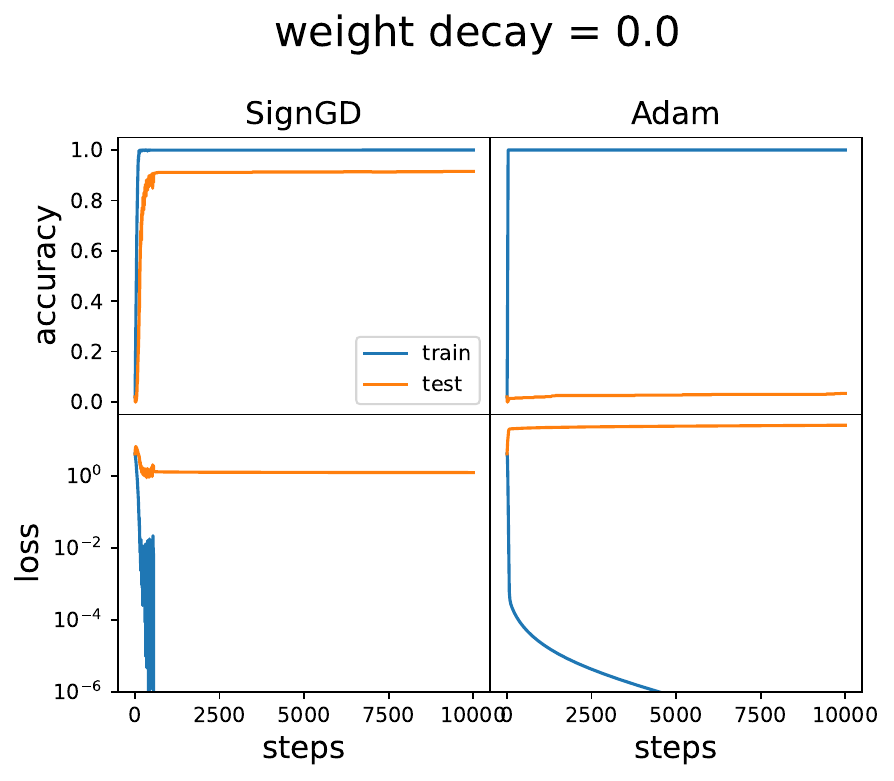}
  \caption{}
  \label{fig:grokking_sub1}
\end{subfigure}%
\begin{subfigure}{0.49\textwidth}
  \centering
  \includegraphics[width=1.0\linewidth]{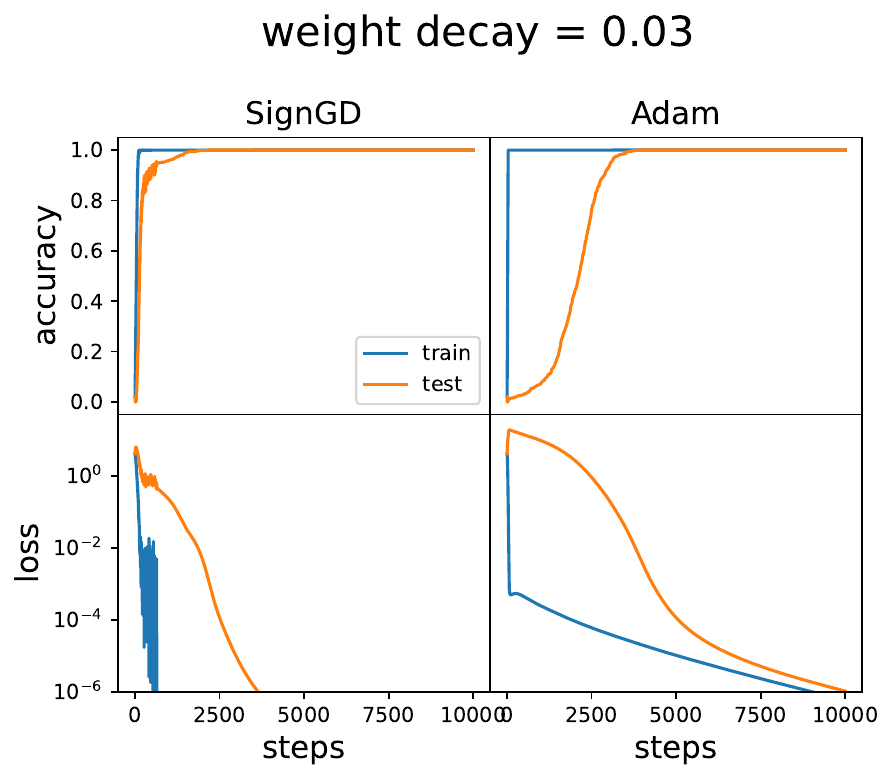}
  \caption{}
  \label{fig:grokking_sub2}
\end{subfigure}
\caption{Grokking. SignGD shows much faster generalization than Adam, almost eliminating grokking.}
\label{fig:grokking}
\end{figure}

{\bf Problem setup} Power et al.~\cite{power2022grokking} observe grokking in learning “algorithmic” binary operations.
Given some binary operation $\circ$, a network is tasked with learning the map $(a, b)\to c$ where $c = a \circ b$. We choose the modular addition problem $a+b\equiv c\ ({\rm mod}\ p), p=59$.
Each token $a$ is represented as a 32-dimensional
embedding vector $\bm{E}_a$. The embeddings are learnable and initialized randomly. A three-layer MLP (100 hidden neurons) takes  $[\bm{E}_a,\bm{E}_b]$ as inputs, and the  
final linear layer maps the output to class logits for each token. We carry out a 0.8/0.2 train/test split and train the MLP with Adam or SignGD with learning rate $10^{-3}$ for 10000 steps. We try both with weight decay 0.03 and without weight decay. SignGD in fact is a special case of Adam: Adam by default chooses $\beta_1=0.9,\beta_2=0.999$, while SignGD is equivalent to $\beta_1=\beta_2=0$. 

As shown in Figure~\ref{fig:grokking}, SignGD shows better and/or faster generalization than Adam. Without weight decay (left panel), SignGD saturates to around 90\% test accuracy while Adam ends up with around zero test accuracy. With a weight decay 0.03, SignGD generalizes faster than Adam. We hypothesize that the benefit of SignGD comes from its smaller $\beta_2$, which controls how fast Adam's denominator decays. Larger $\beta_2$ leads to slower denominator decay, leading to small learning rates, hence slow learning. We leave the full analysis to future work. The point we want to make is that: simplified models can sometimes provide insights for real-world training, and we can benefit from these simple models because they have lower cost, allow more control, and are easier to understand. 

\subsection{Data Reweighting}\label{subsec:weighting}

As we see in Figure~\ref{fig:nalign}, less frequent skills tend to be learned later in training, which becomes worse when $\alpha$ is large. Can we effectively reduce $\alpha$ through re-weighting? The natural idea is that: when a data point from a less frequent skill is seen by the network, a larger weight is placed on it to counteract its rareness. Although this sounds plausible in theory, it is not immediately implementable in practice. We also do not know how to decompose a complex task, say language modeling, into multiple skills, let alone obtain statistics about their frequencies. We adopt a workaround: the sequential learning of skills tells us that a less frequent skill always has higher losses than a more frequent skill, given a fixed training step. This means that we could use losses as a surrogate for frequencies (higher losses mean lower frequencies). So instead of minimizing the standard mean loss $\ell = \frac{1}{B}(\sum_{i=1}^B \ell_i)$, we minimize the weighted mean loss $\ell' = \frac{1}{B}(\sum_{i=1}^B w_i\ell_i)$ where we choose $w_i=\ell_i$, which puts more weight on data points with higher losses. The reweighting itself may not be very novel (e.g., the focal loss~\cite{lin2017focal} that puts more focus on hard examples is similar to our proposal here), but its interpretation in the context of skill learning is new, to the best of our knowledge.

\begin{figure}
    \centering
    \includegraphics[width=1.0\linewidth]{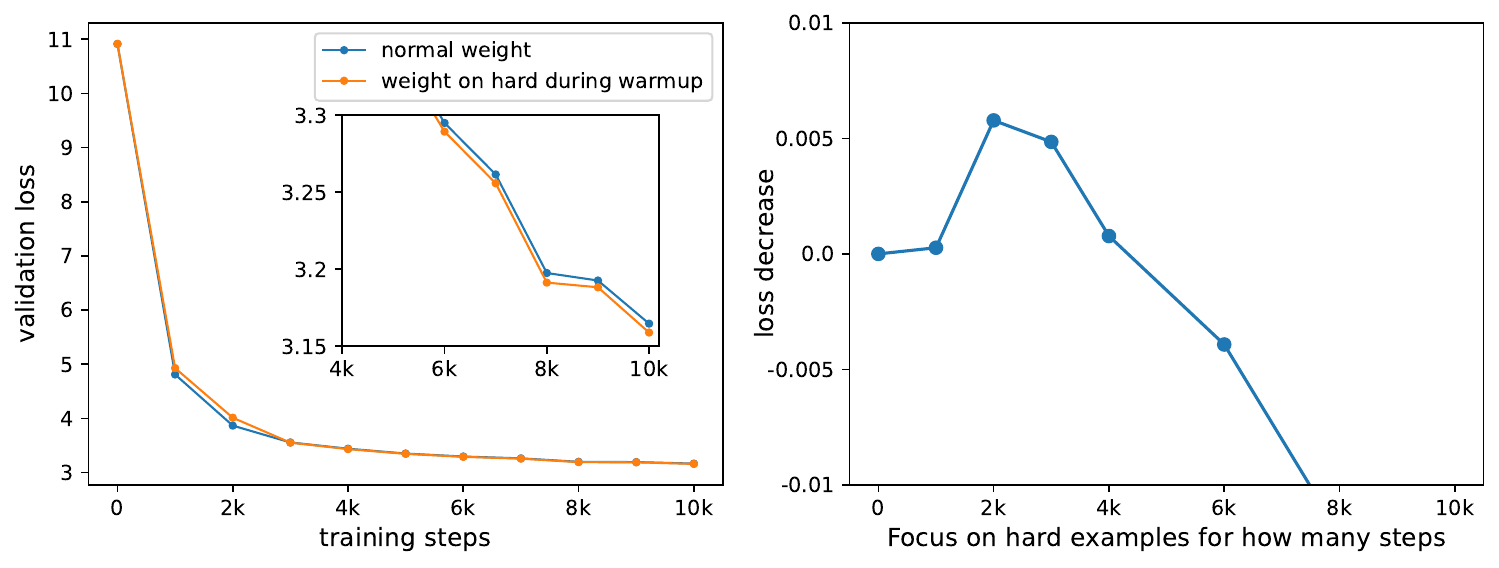}
    \caption{Token reweighting (focusing on hard tokens) for GPT-2 small. Left: learning curves (blue: baseline, uniform weighting; orange: weighting on hard tokens). Right: loss improvement as a function of $A$, the number of steps that focus on hard tokens in the early stage of training.}
    \label{fig:llm-weighting}
\end{figure}

We pre-train a GPT-2 small model (based on NanoGPT) on OpenWebText. We use 8 V100 GPUs, choose block size 1024, batch size 480 blocks. We use the Adam Optimizer, with a linear warmup learning rate schedule for 2k steps to maximum learning rate $6\times 10^{-4}$, and a cosine decay schedule from 2k to 10k, ending at lr $3\times 10^{-5}$. We use the standard mean loss as the baseline. For the weighted loss experiments, the first $A$ steps use the weighted loss while the rest $10{\rm k}-A$ steps use the standard loss. We compute the decrease of validation loss for the weighting runs compared to the baseline, and show them in Figure~\ref{fig:llm-weighting} right. We find that focusing on hard examples for the first 2k steps yields the most speedup (learning curves shown in Figure~\ref{fig:llm-weighting} left), while further increasing weighting steps can reduce improvement or even slow down training. We hypothesize that this is because of noisy tokens~\cite{lin2024rho}. These tokens always have high losses throughout training, so putting large weights on them (especially in the later stage of training) effectively increases noise hence preventing convergence. This suggests that better weighting schemes should take both skill quality and frequency into consideration -- we want to speed up low-frequency, high-quality skills. However, noisy tokens correspond to low-frequency (because noises are very diverse), and low-quality skills. We cannot distinguish between these two solely based on frequencies (losses).

A promising future direction would be inferring frequencies $p$ based on losses $\ell$. In fact, if we assume skill independence and MSE loss, the effective model suggests that $\ell^{1/(2p_i)}=C$:  according to the conservation laws of Eq.~(\ref{eq:eff}), we know $u^{1/p}=C$ for all skills (tokens), and $\ell\equiv u^2$, we have $\ell^{1/(2p_i)}=C$. However, questions remain about how to estimate $C$. Also, conservation law analysis needs to be done for more realistic setups (with skill correlation and dependence, and the cross-entropy loss).

\subsection{Understanding of recently proposed   optimizers}\label{subsec:optimizer}


\begin{figure}[ht]
    \centering
    \includegraphics[width=0.8\linewidth]{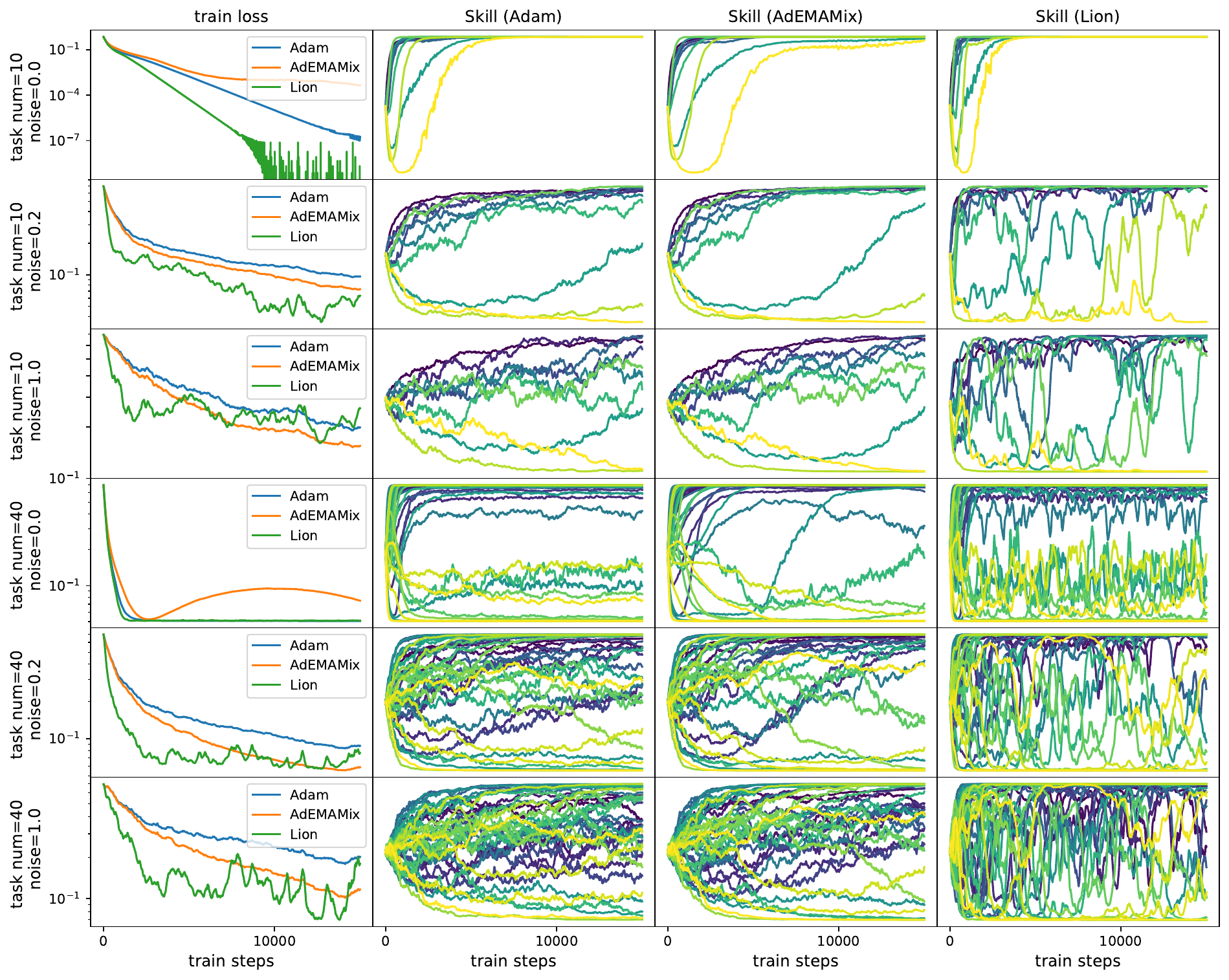}
    \vskip 0.5cm \includegraphics[width=0.8\linewidth]{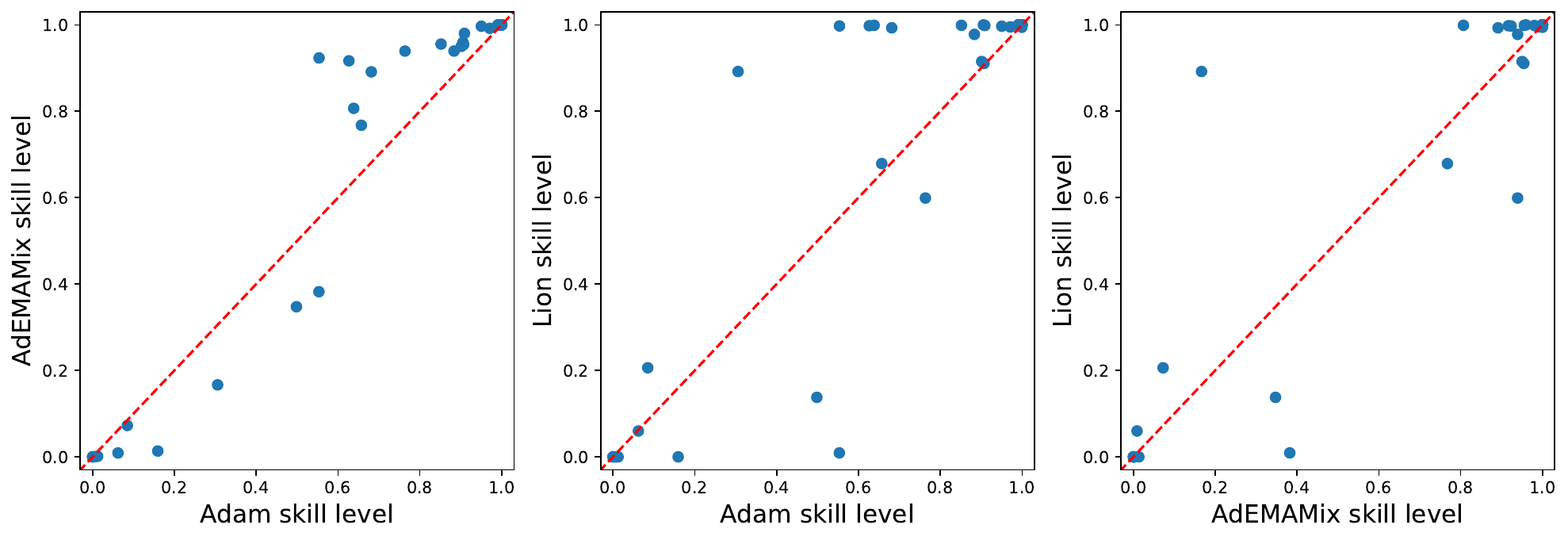}
    \caption{Understanding recently proposed optimizers AdEMAMix and Lion and compare them with Adam. Top: learning dynamics for different $n_{\rm task}$ and gradient noise scale $\sigma$. Each row corresponds to a different $(n_{\rm task},\sigma)$ combination; the first column shows training loss dynamics; the rest columns show skill dynamics with each optimizer. Bottom: AdEMAMix and Lion learn frequent skills better than Adam.  This is for $n_{\rm task}=40$, $\sigma=0.2$ at training step 15000.
    }
    \label{fig:optimizer}
\end{figure}

Pre-training a large language model is very time and energy-consuming. To mitigate this, there is growing interest in the optimizer community in proposing more effective optimizers that can speed up large language model (LLM) training, e.g., Lion~\cite{chen2024symbolic} (discovered by symbolic search), AdEMAMix~\cite{pagliardini2024ademamix} (maintaining an older momentum), Sophia~\cite{liu2023sophia} (combining Adam and Hessian method), Soap~\cite{vyas2024soap} (combining Shampoo and Adam), MARS~\cite{yuan2024mars} (variance reduction), Muon~\cite{jordan2024muon} (adaptive updates in eigenspace). Although these new optimizers show effectiveness in specific setups, why and when they are effective remains unclear. The insights we gained from common ``toy landscapes'' (e.g., convex functions, the Rosenbrock banana function) fail to capture many aspects of real training dynamics (e.g., high dimensionality, batch size effects, etc). Therefore, it is desirable to have a model of intermediate complexity that can model key aspects of LLM training but also remains sufficiently simple and intuitive. We believe that the Geometry model proposed in Section~\ref{sec:models} can nicely serve this purpose. 

We hypothesize that the biggest reason for the slow training dynamics of LLM is due to these two patterns in skill interaction: (1) Correlation (Interference). LLMs are under-parametrized, meaning skills should be correlated with each other in order to be packed into the small parameter space. Some skills could have negative interference (conflicting objectives), making convergence hard. (2) Many skills are very sparse. These skills do not occur much in data, so a good optimizer should be sensitive to these rare skills and accumulate their effects, instead of ignoring them or treating them as noises. In this mental picture, both underparmatrization ($n_{\rm task}>n_{\rm dim}$) and noise should play big roles, which we vary in our Geometry model to study their effect. Fix $n_{\rm dim}=10$, we visualize the training dynamics of losses and skills in Figure~\ref{fig:optimizer} top panel, under different conditions $n_{\rm task}=\{10,40\}$ and $\sigma=\{0,0.2,1.0\}$, with three optimizers -- Adam~\cite{kingma2014adam}, AdEMAMix~\cite{pagliardini2024ademamix} and Lion~\cite{chen2024symbolic}. We choose $\alpha=2$ and cross-entropy loss.  

{\bf AdEMAMix}~\cite{pagliardini2024ademamix} maintains a slow momentum with $\beta_3=0.9999$ in addition to a quick-varying momentum with $\beta_1=0.9$ as in Adam. The idea is that the slow momentum is more robust with noises since it averages over more steps. However, the slow momentum may introduce systematic error in fast-changing landscapes. To address this, the authors wrote -- ``While changing the direction of the slow momentum is difficult, any adjustment orthogonal
to that direction is easy—which favors fast progress in sinuous canyon-like landscapes''. In our Geometry model, AdEMAMix shows speed up when there are large gradient noises. The advantage becomes even more significant when $n_{\rm task}$ becomes larger. This agrees with our intuition: (1) the slow momentum, averaging across many steps, can reduce the variance of the noises. (2) the effect of sparse skills can also be accumulated in the slow momentum, reaching equilibrium faster. However, AdEMAMix seems to have slower convergence or oscillatory behavior than Adam when gradient noises are small, which is not too surprising because the slow momentum has the issue of overshooting and slow adaptation. In large noise settings, the advantages (averaging out  noises, accurate estimation of skill frequency) outweigh the drawbacks. More mechanistically, we can understand the advantages and drawbacks from skill dynamics. Shown in Figure~\ref{fig:optimizer} top panel 1st row 3rd column, the yellow curve (the least frequent skill) shows slower convergence; Shown in 4th row 3rd column, skill dynamics display overshooting (not converge yet in given training steps). While in 2nd row 3rd column or 5th row 3rd column, it is clear that AdEMAMix induces skill dynamics converging faster than Adam. 

{\bf Lion}~\cite{chen2024symbolic} is an optimizer discovered by symbolic search. Lion converges fast initially but displays sudden jumps in training or large oscillations around convergence. Interestingly, we can reproduce the same phenomenon in the Geometry model. As shown in Figure~\ref{fig:optimizer} first column, Lion has the largest converging rate initially almost in all cases, however, learning becomes slower and dominated by fluctuations at later stages (when noises are large). In the last column, we see a similar oscillation in skill dynamics -- skills move fast initially (quickly deciding whether being learned or not), but display huge fluctuations afterward. This suggests that Lion can make quick movement, which explains its superior initial convergence. However, the random fluctuations at the later training stage may prohibit further convergence, especially when the gradient noise (small batch size) is very large. 

Different optimizers can produce models with different ``personalities'' -- some prefer exploitation (master less frequent skills perfectly while giving up on frequent skills), and some prefer exploration (learn almost all skills but learn every skill only to a partial level). Some are more organized (learn frequent skills first), while some seem more chaotic (learn infrequent skills while failing frequent ones). We compare skill levels for the three optimizers in Figure~\ref{fig:optimizer} (bottom). When comparing Adam and AdEMAMix, AdEMAMix has higher skill levels than Adam for those skills that Adam mastered above average, whereas has lower skill levels than Adam for those that Adam mastered below average. AdEMAMix exploits more than Adam by prioritizing more frequent skills, explaining AdEMAMix's lower loss. When comparing Adam and Lion, Lion pushes many skills to extremes: either completely learning or completely failing, and the decision on which skill to learn or not is somewhat random: some frequent skills are not learned, while some less frequent skills are learned.

\section{Implications for task compositionality}
\label{sec:dependent}


Before we move on, let us reflect on our philosophy: In Section~\ref{sec:models}, we proposed a Geometry model and then proposed the simpler Resource model that mimics the behavior of the Geometry model. The proposal of the Resource model may come off as being redundant and pointless -- because one can simply fully simulate the Geometry model without ever resorting to the Resource model. However, Resource models can still be useful because (1) when a Geometry model is available, Resource models are usually simpler than the Geometry model, providing more intuition and insights. (2) it is sometimes the case that Geometry models are not available or too cumbersome to construct, while it is macroscopic behavior (rather than microscopic details) that is of interest. The goal in this section (modeling dependent skills) falls into the second category, where it might be too cumbersome to specify the details about gradients in the parameter space to model skill dependence. Instead, Resource models are much easier to model task dependence. 

\subsection{Dependent skills can also lead to the Domino effect}

\begin{figure}
    \centering
    \includegraphics[width=1.0\linewidth]{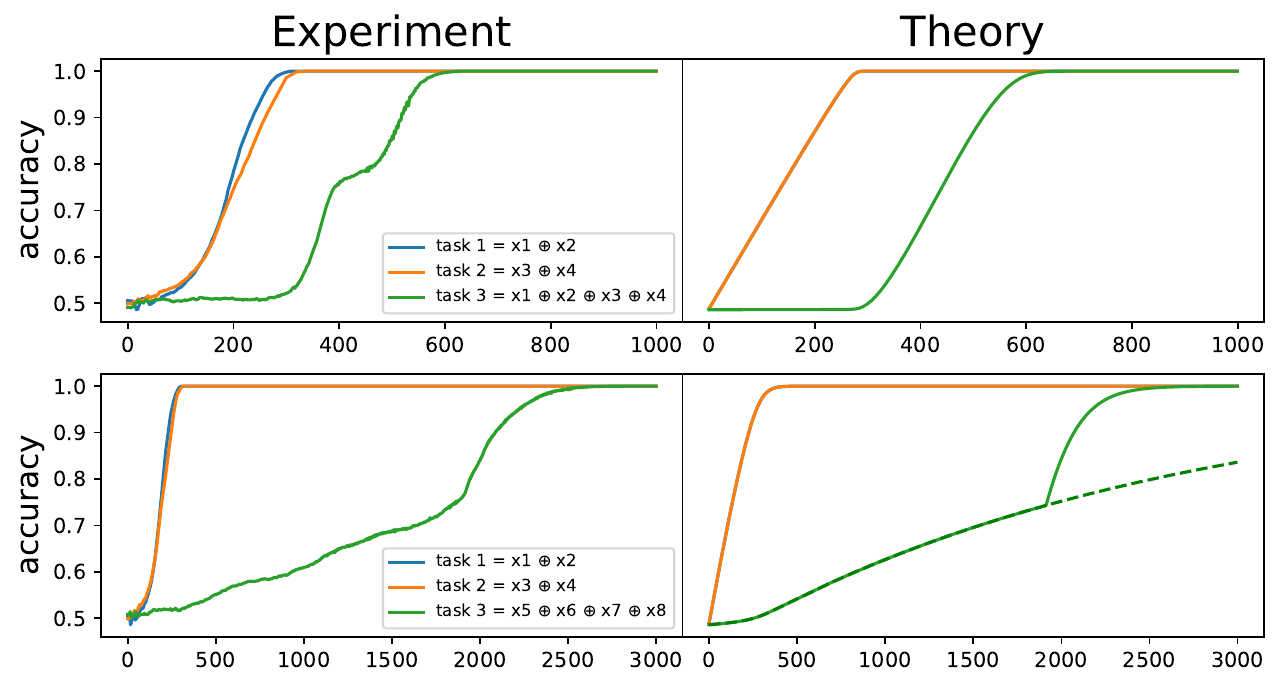}
    \caption{MLP experiments on sparse parities. Top: The third task depends on the first two. Bottom: The third task does not depend on the first two. Our Resource model can capture key features of learning dynamics in both cases.}
    \label{fig:sp_eff_exp_compare}
\end{figure}

Suppose a neural network takes in a bit string $(x_1, x_2, \cdots, x_n)\ (n=32)$ and is tasked with computing three sparse parities $y_1=x_1\oplus x_2$, $y_1=x_3\oplus x_4$, $y_3=x_1\oplus x_2\oplus x_3\oplus x_4$. It is known that learning $k$-parity ($k$ is the number of bits to be summed) becomes exponentially slow when $k$ increases~\cite{shalev2017failures}. Consequently, a good strategy for the network would be computing $y_3$ not from scratch, but instead, learning $y_1$ and $y_2$ first, and then leveraging $y_3=y_1\oplus y_2$ to compute $y_3$. If this is the case, we expect to see the learning of task 3 only begins after both tasks 1 and 2 complete learning. This is indeed the case empirically, as shown in Figure~\ref{fig:sp_two_mechs} right or Figure~\ref{fig:sp_eff_exp_compare} top left. We train a two-layer MLP (50 hidden neurons) on 10000 training samples (full batch) using the Adam optimizer (learning rate $10^{-3}$).

Intriguingly, it appears that task 3 kicks off learning right after the first two tasks get perfect accuracy. There are two hypotheses for this to happen. {\bf Hypothesis 1:} learning task 3 is independent of tasks 1 and 2. Although all three tasks have the same frequencies, task 3 has smaller gradients initially due to larger $k$, so this comes back to the resource explanation in the last section. {\bf Hypothesis 2:} learning task 3 is dependent on task 1 and 2. To rule out the first hypothesis, we conduct an ablation experiment by replacing the third task with $y_3'=x_5\oplus x_6\oplus x_7\oplus x_8$, which is independent of the first two tasks, whose skill dynamics is shown in Figure~\ref{fig:sp_eff_exp_compare} bottom left. It is clear that learning of $y_3'$ compared to $y_3$ is much delayed, suggesting that the network must have leveraged (at least in some implicit way) $y_1$ and $y_2$ to compute $y_3$. 

{\bf Resource model} Can we induce similar skill dynamics  (Figure~\ref{fig:sp_eff_exp_compare} top left) using Resource models? It might be cumbersome to deal with the Geometry model since specific details must be filled in (how does ``task dependence'' in the language of gradients?) to complete the Geometry model. Instead, it is more convenient to work with the Resource model. We can add a small modification to Eq.~(\ref{eq:eff}) (we have set $p_1=p_2=p_3=1$):

\begin{align}\label{eq:eff_dep}
\frac{du_i}{dt} & = -\eta_{\rm eff}\frac{u_i}{u_1+u_2+u_3+N_0},\quad i=1,2 \\
\frac{du_3}{dt} & = -\eta_{\rm eff}\frac{{\color{red} B(u_1,u_2)}u_3}{u_1+u_2+u_3+N_0}. \\
\end{align}
where $B(u_1,u_2)$ is supposed to be large only when both $u_1$ or $u_2$ are small, and we choose $B(u_1,u_2)=(1-u_1)^\gamma (1-u_2)^\gamma$ with $\gamma=0.01$, which can be understood as a soft version of the AND operation. The resulting dynamics is shown in Figure~\ref{fig:sp_eff_exp_compare} top right, showing decent agreement with experiments. 

When learning $(y_1,y_2,y_3')$, since all three tasks are independent, we can use Eq.~(\ref{eq:eff}) to simulate skill dynamics, which is shown in Figure~\ref{fig:sp_eff_exp_compare} bottom. Note that if we use a time-independent $p_3=0.0045$, the learning curve (dashed green) for task 3 only agrees with the first 2000 steps. However, if we phenomenological set $p_3$ to be larger after 2000 steps $p_3=0.08$, we can get a faster increase (solid green) which agrees better with experiments. This makes sense because when learning parities, gradients are small at random initialization but become larger after getting near a basin of attraction.

\subsection{Simulating skill dynamics with arbitrary dependence graphs}

\begin{figure}
    \centering
    \includegraphics[width=1.0\linewidth]{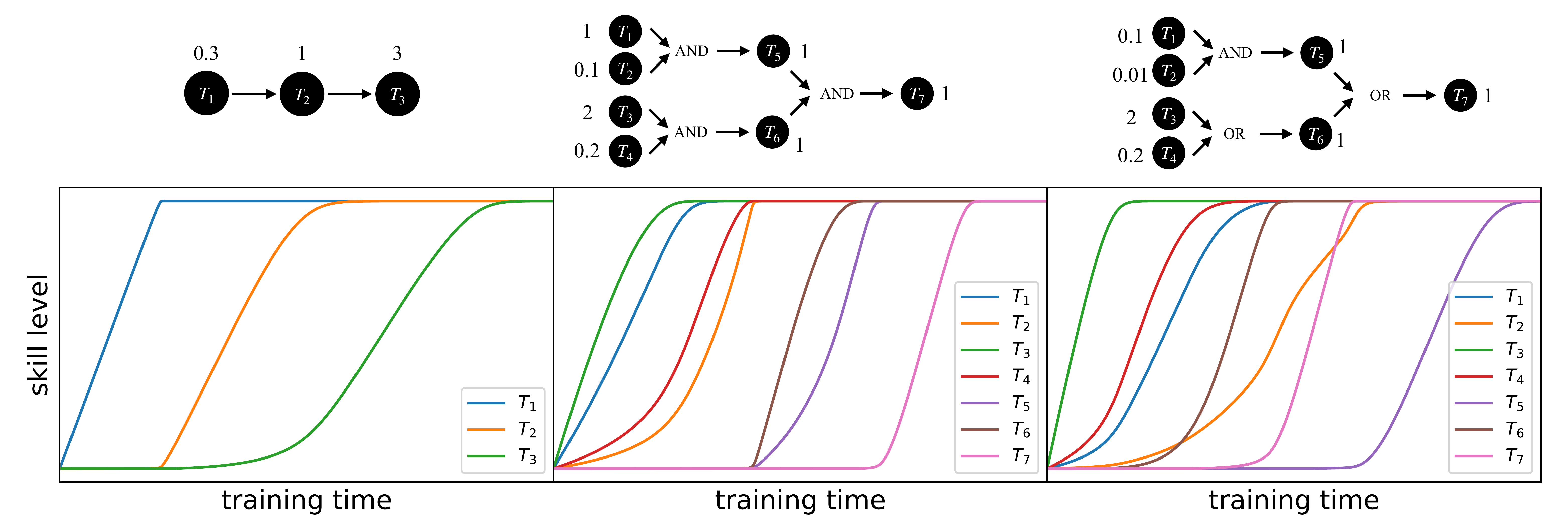}
    \caption{The Resource model can produce insightful skill dynamics, for any task dependency graphs.}
    \label{fig:eff_hier}
\end{figure}

With the Resource model, we are now able to simulate arbitrarily complex skill dynamics, as long as skill frequencies and the dependence graph are given. For the task $i$, its dependence on other skills can be effectively modeled as a phenomenological function $B_i(u_1,\cdots, u_{n_{\rm task}})$:
\begin{equation}\label{eq:eff_graph}
    \frac{du_i}{dt} =-\eta_{\rm eff}\frac{{\color{red} B_i(u_1,\cdots,u_{n_{\rm task}})}p_iu_i}{(\sum_{j=1}^{n_{\rm task}} p_ju_j)+N_0}.
\end{equation}
For example, if skill $k$ starts to learn only when both $i$ AND $j$ completed learning, we can set $B_k(u_i,u_j)=(1-u_i)^{\gamma}(1-u_j)^\gamma$, as we did above. If skill $k$ starts to learn when either $i$ OR $j$ completed learning, we can set $B_k(u_i,u_j)={\rm max}((1-u_i)^{\gamma},(1-u_j)^\gamma)$, which is a soft version of the OR operation. For other dependence relations, we can define corresponding $B$ functions to model them. We show in Figure~\ref{fig:eff_hier} that our Resource model is able to induce intriguing dynamics that agree with specifications about task dependence. In the first case, task 3 is triggered by task 2, which in turn is triggered by task 1. Although task 1 has the lowest frequency while task 3 has the highest frequency, the sequence goes as $1\to 2\to 3$ because of the dependence structure. The second case showcases a hierarchical dependence graph. The thrid case is similar to the second case except that two ANDs are replaced by ORs. This induces interesting dynamics where task $7$ is not the last one to finish learning even though it is a descendant of all other tasks. The learning curve of task 2 is also interesting: demonstrating several speed-ups and slow-downs, corresponding to the beginning or completion of other skills (occupying or freeing resources).

To be clear, we do not expect that our Resource model can capture all details of skill dynamics in practice, but it provides a decent reference that may already explain many seemingly complicated phenomena. The ultimate goal is the \textit{inverse problem} -- infer the skill dependency graph and skill frequencies from training dynamics. This is no doubt an extremely challenging problem. Our Resource model takes a humble first step by providing a solution to the \textit{forward problem} -- use skill frequencies and dependency graph to calculate training dynamics.

\section{Implications for Modularity}\label{sec:modularity}

\begin{figure}
\centering
\begin{subfigure}{.55\textwidth}
  \centering
  \includegraphics[width=1.0\linewidth]{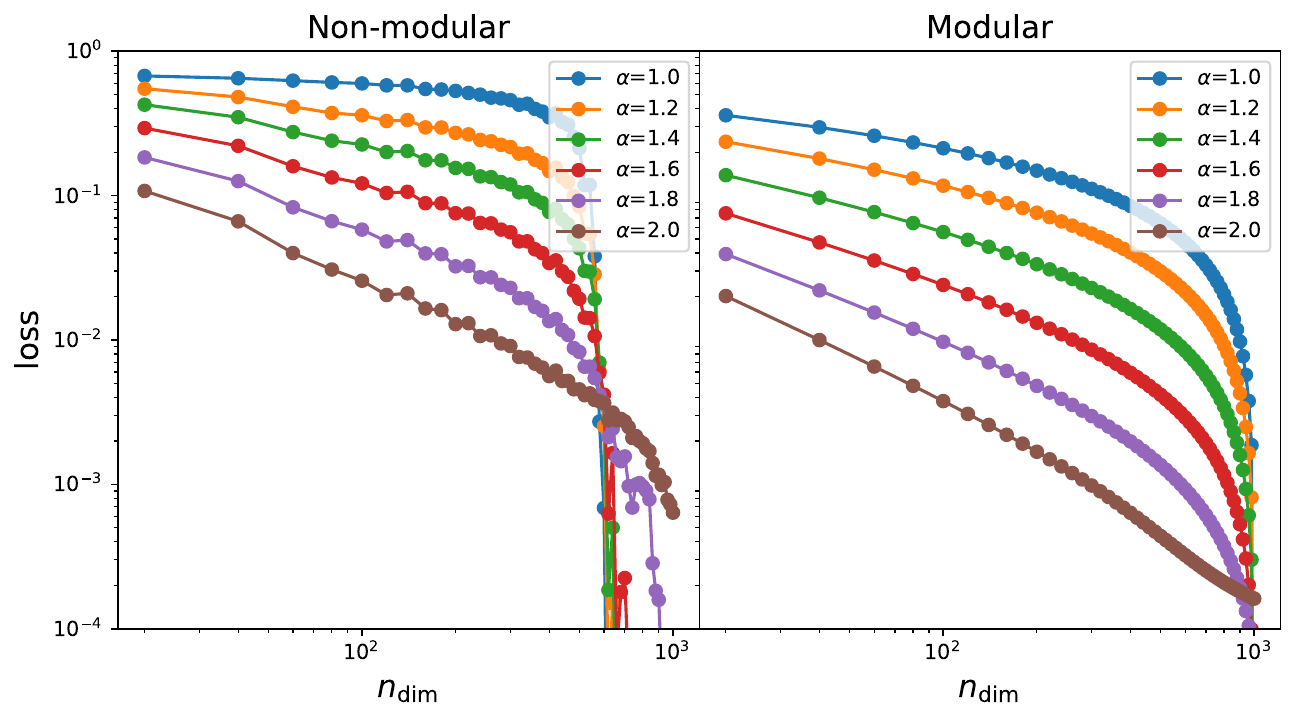}
  \caption{}
  \label{fig:sub1}
\end{subfigure}%
\begin{subfigure}{0.38\textwidth}
  \centering
  \includegraphics[width=1.0\linewidth]{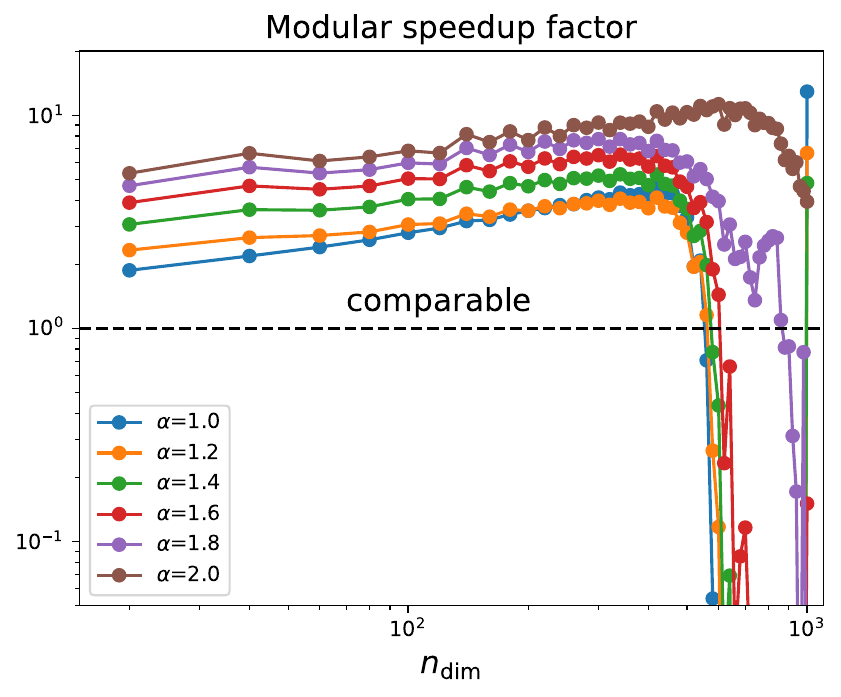}
  \caption{}
  \label{fig:sub2}
\end{subfigure}
\begin{subfigure}{.55\textwidth}
  \centering
  \includegraphics[width=1.0\linewidth]{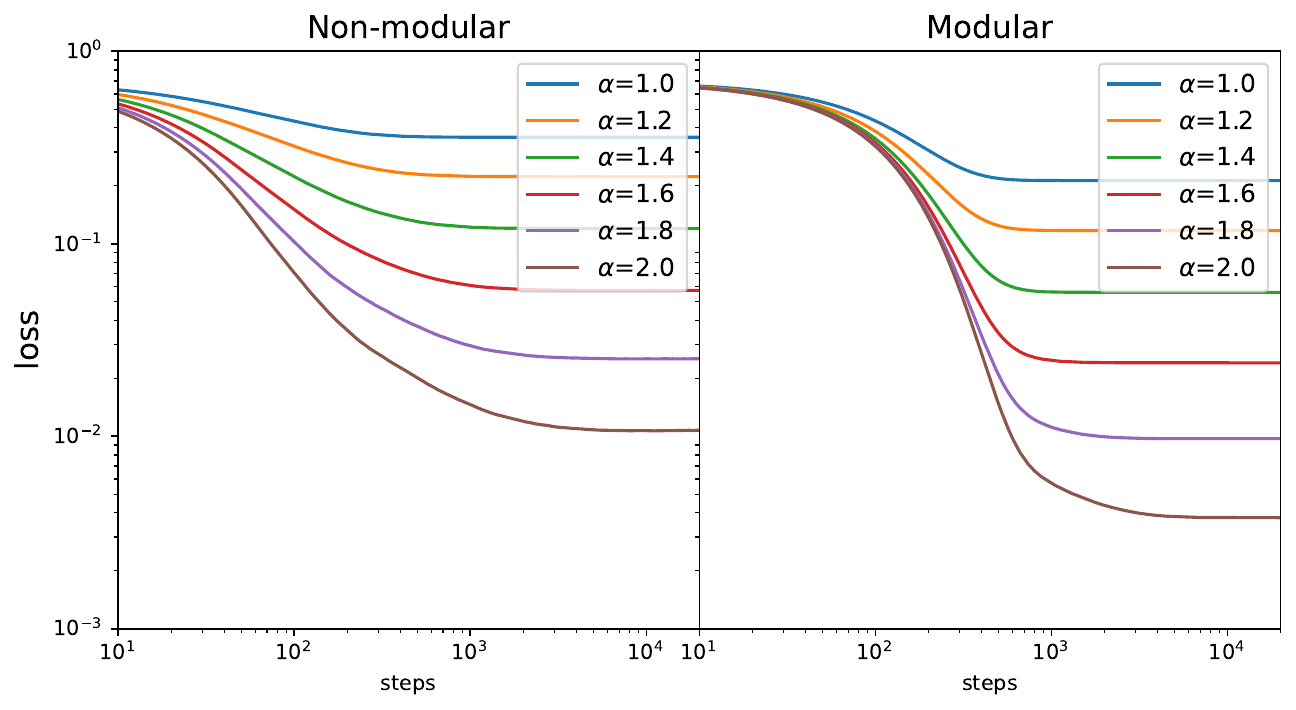}
  \caption{}
  \label{fig:sub3}
\end{subfigure}%
\begin{subfigure}{0.38\textwidth}
  \centering
  \includegraphics[width=1.0\linewidth]{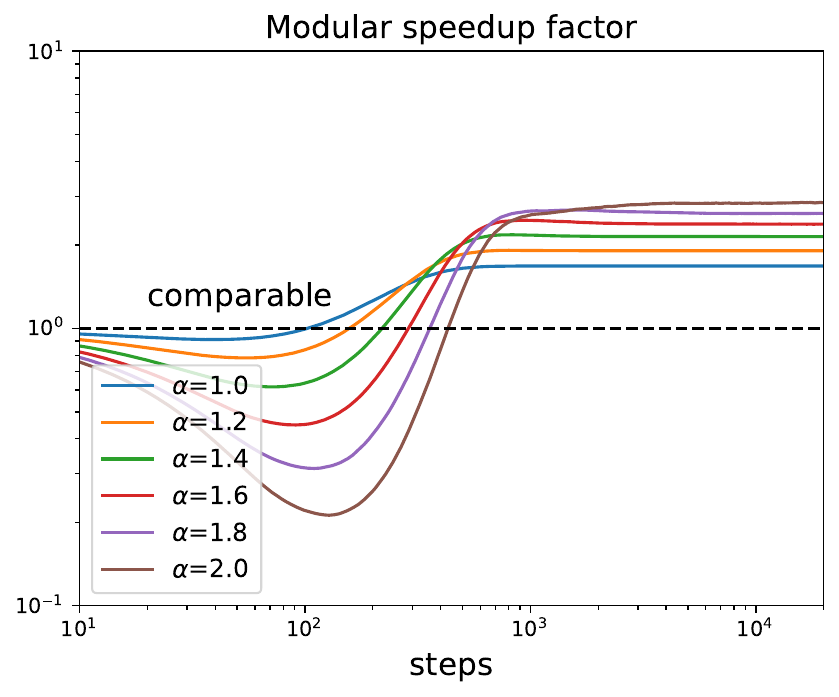}
  \caption{}
  \label{fig:sub4}
\end{subfigure}
\caption{We demonstrate the benefits of modularity using the meachanistic model. (a) loss against $n_{\rm dim}$ for modular and non-modular models. (b) Modular speedup factor (non-modular loss over modular loss) against $n_{\rm dim}$; (c) loss against $S$ (training steps) for modular and non-modular models. (d) Modular speedup factor (non-modular loss over modular loss) against $S$. }
\label{fig:mod-speedup}
\end{figure}

\begin{figure}
    \centering
    \includegraphics[width=0.8\linewidth]{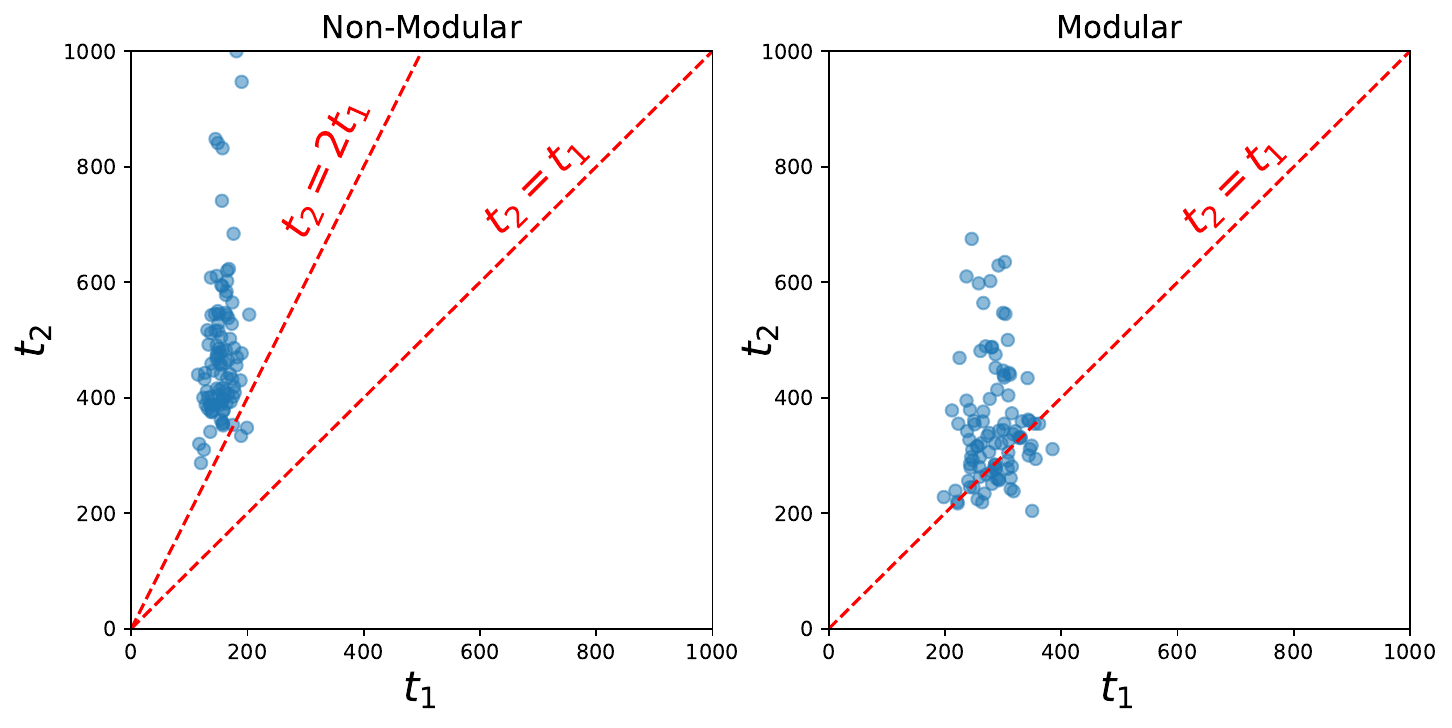}
    \caption{Benefits of modularity for imbalanced tasks. Training a standard (non-modular, left) and modular (right) MLPs for the regression task $(x,y)\to (x^2,y^2)$ where $x$ is dense while $y$ is sparse with sparsity $0.01$. Each task is considered successful when MSE < 0.001 and the successful time is denoted $t_1$ (for $x$) and $t_2$ (for $y$) respectively. Each scatter corresponds to a random seed, and we run 100 random seeds in total. Non-modular networks demonstrate clear sequential learning, while modular networks do not hence speed up by leveraging ``parallelism''. }
    \label{fig:modularity}
\end{figure}

As we learned from the Domino model in Section~\ref{subsec:domino}, some skill takes a long time to learn simply because its frequency is outweighed by other skills. This story is only valid when different skills interfere (their gradients share the same parameter space). What if we partition the parameter space into a few disentangled parts, and each part is responsible for one skill? We call this modularity, and assess the benefits (if any) of modular models through the perspective of the Domino model. 

\subsection{Insights from the Domino model}

The Domino model considers the extreme case where we have $n_{\rm task}$ tasks whose frequencies are $p_1\gg p_2\gg \cdots\gg p_{n_{\rm task}}$. This hierarchy would lead to strict sequential learning of skills, i.e., skill $i+1$ starts to learn only after skill $i$ completes learning. When a skill is being learned, it takes all the resources $n_{\rm dim}$ which has an effective learning rate $\eta_{\rm eff}\propto \sqrt{n_{\rm dim}}$ (Eq.~(\ref{eq:eff})). This means that each task takes time $t_0\propto \eta_{\rm eff}^{-1}\propto 1/\sqrt{n_{\rm dim}}$ to learn, and the total learning time for all tasks is $T_{N} = n_{\rm task}t_0\propto n_{\rm task}/\sqrt{n_{\rm dim}}$ (subscript N for Non-modular). Now we consider a modular model where $n_{\rm dim}$ dimensions are equally divided into $n_{\rm task}$ parts, so each skill can consume only $n'_{\rm dim}=n_{\rm dim}/n_{\rm task}$ dimensions but these dimensions are dedicated, i.e., not being shared by other skills. The effective learning rate is then $\eta'_{\rm eff}\propto \sqrt{n'_{\rm dim}}=\frac{\sqrt{n_{\rm dim}}}{\sqrt{n_{\rm task}}}$, meaning that it takes time $t_0'\propto 1/\eta_{\rm eff}^{-1}\propto \sqrt{n_{\rm task}}/\sqrt{n_{\rm dim}}$ to learn each skill. Since all skills are learned in parallel in their own dimensions, the total learning time $T_M=t_0'\propto \sqrt{n_{\rm task}}/\sqrt{n_{\rm dim}}$ (subscript M for modular). Now we compare $T_N$ and $T_M$ by noticing their scaling with $n_{\rm task}$: $T_N\propto n_{\rm task}$, $T_M\propto \sqrt{n_{\rm task}}$, which means the modular networks learn faster than non-modular networks. We want to make a caveat: although this speed-up is plausible in theory, it may not be very useful in practice since we do not know a priori how to decompose a complex task into skills. When we do know how to decompose skills, the theoretical advantage is achievable, as we show in the proof-of-concept examples below.

{\bf Modular Parallelism vs Quantum Parallelism} Let us make a superficial analogy with quantum parallelism. The Grover's algorithm, which is a quantum algorithm for search (find one thing out of $N$ items), can reduce the search time from $O(N)$ to $O(\sqrt{N})$ thanks to quantum parallelism. In our case, it is coincident to notice that a modular network can also reduce learning time from $O(n_{\rm task})$ to $O(\sqrt{n_{\rm task}})$ by leveraging parallelism enabled by modularity.

\subsection{Insights from the Geometry model}
A standard non-modular model randomly samples task vectors which spread across all dimensions, while a modular model has one-hot task vectors such that each task is ``localized'' in one parameter. We stick to the setup in Section~\ref{sec:nsl} ($n_{\rm task}=1000$). We compare the loss scaling against $n_{\rm dim}$ (number of parameters) and $S$ (training steps) in Figure~\ref{fig:mod-speedup} (a) and (c). To clearly show the benefits (or disadvantages) of modular models, we compute the ratio of the non-modular loss over the modular loss and plot them in (b) and (d). Modular models have advantages when: (1) $n_{\rm dim}$ is not too big to be near the critical point. (2) in the later stage of training. In fact, modular models have higher losses in the initial stage of training, because non-modular networks put most resources into learning frequent skills (leading to a significant loss decrease), while modular networks allocate equal resources to all skills such that all skills are learned at a similar pace. This synchronization feature makes modular models advantageous in the later stage of training when non-modular models still struggle to learn less frequent skills.

\subsection{"Real" MLP experiments}
The task is to take in two real numbers $(x,y)$ and predict their squared values $(x^2,y^2)$. We manually create a hierarchy in training data by imposing sparsity to $y$, i.e., $y$ is zero 99\% of the time, while randomly drawn from $U[-1,1]$ 1\% of the time. By contrast, $x$ is always drawn from $U[-1,1]$. We train a [2,200,200,2] MLP using the Adam optimizer (learning rate $5\times 10^{-4}$) with the MSE loss on 1000 data points. We count a task as successfully learned when the MSE loss is smaller than 0.001. Because of the sparsity of $y$, we expect that $y$ is learned more slowly due to smaller gradients, which is indeed the case as we show in Figure~\ref{fig:modularity} left. In particular, the time to learn the second task $t_2$ seems to be lower bounded by $2t_1$ for almost all 100 random seeds ($t_1$ is the time to learn the first task), which agrees with the observation of the Domino effect. If we know the modularity structure of the task: the first label $x^2$ only depends on the first input $x$, while the second label $y^2$ only depends on the second input $y$, we can split the MLP into two subnetworks which do not have cross connections. This is equivalent to two [1,100,100,1] MLPs being trained independently on $x$ and $y$, respectively. As shown in Figure~\ref{fig:modularity} right, $t_2\approx t_1$ for most random seeds although $t_2$ has a slightly heavier tail potentially because of the sparsity of $y$ itself.  

Modular neural networks have been around for long~\cite{pfeiffer2023modular,andreas2016neural,alet2018modular}, but they have regained attention recently because of their success in language modeling (the mixture-of-expert model\cite{masoudnia2014mixture, cai2024survey} is a form of modular models). Their benefits are mostly understood from task compositionality, module reuse, ensembling, etc. Our analysis provides another benefit of modularity in neural networks, i.e., speed up training. However, it is practically challenging to decide how to partition the networks into modules. A recent work on gradient routing~\cite{cloud2024gradient},  which applies data-dependent, weighted masks to
gradients during backpropagation, seems a promising approach but has not been scaled up to language models yet.

\section{Related Works}

{\bf Learning Theory} Understanding the success of neural networks (or failure thereof) is a key goal of learning theories. Mainstream learning theories -- one is PAC learning~\cite{mohri2018foundations}, based on computational theory; another family of methods is based on statistical physics~\cite{engel2001statistical,bahri2020statistical}, including replica theory and mean field theory (neural tangent kernel~\cite{jacot2018neural}). These ``classic'' theories are (partially) successful in characterizing static properties of neural networks, while dynamic properties can only be described by more advanced and complex theories, e.g., dynamical mean field theories. Singular learning theory~\cite{watanabe2007almost,wei2022deep} seems promising to describe training dynamics especially characterize phase transitions, but the computation RLCT is not very tractable to make insights for real-world setups.

Although these classic theories have solid mathematical foundations,  they are sometimes irrelevant for what is observed in practice. They have two limitations: (1) their modeling of data might be overly simplistic (e.g., assuming Gaussian distribution), which fails to model the compositional structure of complex tasks (e.g., language modeling). (2) their modeling of neural networks is too ``mean field'', failing to capture the interaction between different parameters. In practice, it is the collective behavior of parameters (rather than single parameters) that matter, e.g., semantic meanings emerge in some directions in the parameter space, not necessarily aligning with preliminary bases. To put it in physics terms, in many condensed matter systems, the conceptualization of quasi-particles (collective motion of particles) enables much simpler theories than theories based on particles. Reasonable abstractions could make things much simpler.

In the philosophy of abstraction, there is a recent trend of effective theories that treat skills~\cite{arora2023theory, chen2024skill, michaud2024quantization,ren2024towards} as elementary elements and study how their interactions affect learning dynamics. These models can nicely take data structure into consideration, by effectively modeling interactions between skills. However, they either focus on analyzing static properties of neural networks, or only provide qualitative (not quantitative) explanations for learning dynamics, or make too strong assumptions that might not be realistic or generalizable. We complement the literature by translating intuition into mathematical models that can make quantitative predictions while respecting many aspects of real-world learning. There are three previous works that are most related to this paper. The main differences lie in their different abstractions: Arora \& Goyal~\cite{arora2023theory} made a mechanistic model to build the mapping between texts and skills (specific to languages), Chen et al.~\cite{chen2024skill} models skills as a dependency graph that have an optimal learning order, and Michaud et al.~\cite{michaud2024quantization} models skills (they called quanta) as independent with varying frequencies. 
Our model provides a high-level abstraction to task structure since it does not assume the structure of languages as in~\cite{arora2023theory}; this is a double-sided coin -- it is able to provide insights for non-language tasks as well, but it lacks a mechanistic understanding of how language data is mapped to skills. Moreover, our model (the Geometry model) is the only model that has a ``placeholder'' for optimizers (instead of assuming gradient descent or ignoring it), enabling the study of optimizers (as we did in Section~\ref{subsec:optimizer}). Overall, we believe that each model is able to capture some aspects of reality, so promising future directions would be creating new models that can describe aspects not captured by existing models, and combining/unifying existing models to obtain a unified theory (similar to the Standard Model in physics).

{\bf Task Arithmetic} A task vector specifies a direction in the weight space of a pre-trained model, such that movement in that direction improves performance on the task. Task arithmetic refers to the observation that these task
vectors can be modified and combined together through arithmetic operations
such as negation and addition, and the behavior of the resulting model is steered
accordingly~\cite{ilharco2022editing,ortiz2024task,todd2023function}. Our Geometry model relies on the task arithmetic structure and the loss function assumes the weight dis-entanglement condition in~\cite{ortiz2024task}. There are also discussions around how task vectors emerge in in-context learning~\cite{hendel2023context,yang2025taskvectorsincontextlearning}.

{\bf ``Phase transitions'' in neural networks} Complex systems usually demonstrate complex phase diagrams, and neural networks are no exceptions. Phase transitions can happen in the training step, when control parameters are varied, etc. Famous examples of phase transitions in neural networks are the emergence of induction head~\cite{olsson2022context}, grokking~\cite{power2022grokking,liu2022towards,liu2022omnigrok,nanda2023progress}, saddle-saddle leaps~\cite{abbe2023sgd}, emergent abilities~\cite{wei2022emergent}, and algorithms~\cite{zhong2024clock,cui2024phase}, repon collapse~\cite{baek2024geneft}, etc. Initialization scales are shown to be important control parameters that determine stable or divergent of training, and training dynamics is most effective on the edge of chaos~\cite{hayou2019impact, bahri2020statistical} or induces maximal updates~\cite{yang2020feature}. However, there are some debates regarding whether ``phase transitions'' are really sharp or has hidden gradual progress~\cite{schaeffer2024emergent, nanda2023progress}.

{\bf Neural scaling laws} refer to the phenomenon where model performance improvement as resources (data, model parameter, compute) scale up~\cite{kaplan2020scaling,hoffmann2022training}. Neural scaling laws are interesting both in theory and in practice. To understand the origin of neural scaling laws, people proposed theories based on regression on data manifold~\cite{sharma2020neural}, power-law distribution of skills~\cite{michaud2024quantization}, resource interpretation~\cite{song2024resource}, kernel learning~\cite{bahri2024explaining}, random feature modle~\cite{maloney2022solvable}, dynamical mean field theory~\cite{bordelon2024dynamical} etc. There are discussions on how to beat beyond current neural scaling laws, e.g., via data pruning~\cite{sorscher2022beyond}, new architectures (Kolmogorov-Arnold Networks~\cite{liu2024kan}, mixture of expert~\cite{du2022glam}). Researchers are also actively thinking about what are the correct quantities to scale up, e.g., ``think time'' (which can significantly increase language models' reasoning abilities), precision~\cite{kumar2024scaling}, sparsity~\cite{frantar2023scaling}.

{\bf Optimizers} Despite the family of adaptive optimizers (Adam, Adagrad) dominating the deep learning world, there are newly proposed optimizers that are shown to demonstrate superior performance than Adam, especially for language model pretraining. These methods include Lion~\cite{chen2024symbolic}, AdEMAMix~\cite{pagliardini2024ademamix}, MARS~\cite{yuan2024mars}, cautious optimizers~\cite{liang2024cautious}, modular optimizers~\cite{large2024scalable}, Sophia~\cite{liu2023sophia}, soap~\cite{vyas2024soap} (combining Adam with Shampoo~\cite{gupta2018shampoo}), muon~\cite{jordan2024muon}, etc. Despite the effectiveness of these new optimizers on specific setups, it remains mysterious why and when these optimizers are effective (or not). Our Geometry model is a good testbed for optimizers (as we did in Section 6.3), since the model is simple enough to make intuition with, but also is complex enough to capture real-world features (high dimensionality, conceptualization of skills).

\section{Conclusions}

\begin{figure}[ht]
    \centering
    \includegraphics[width=1.0\linewidth]{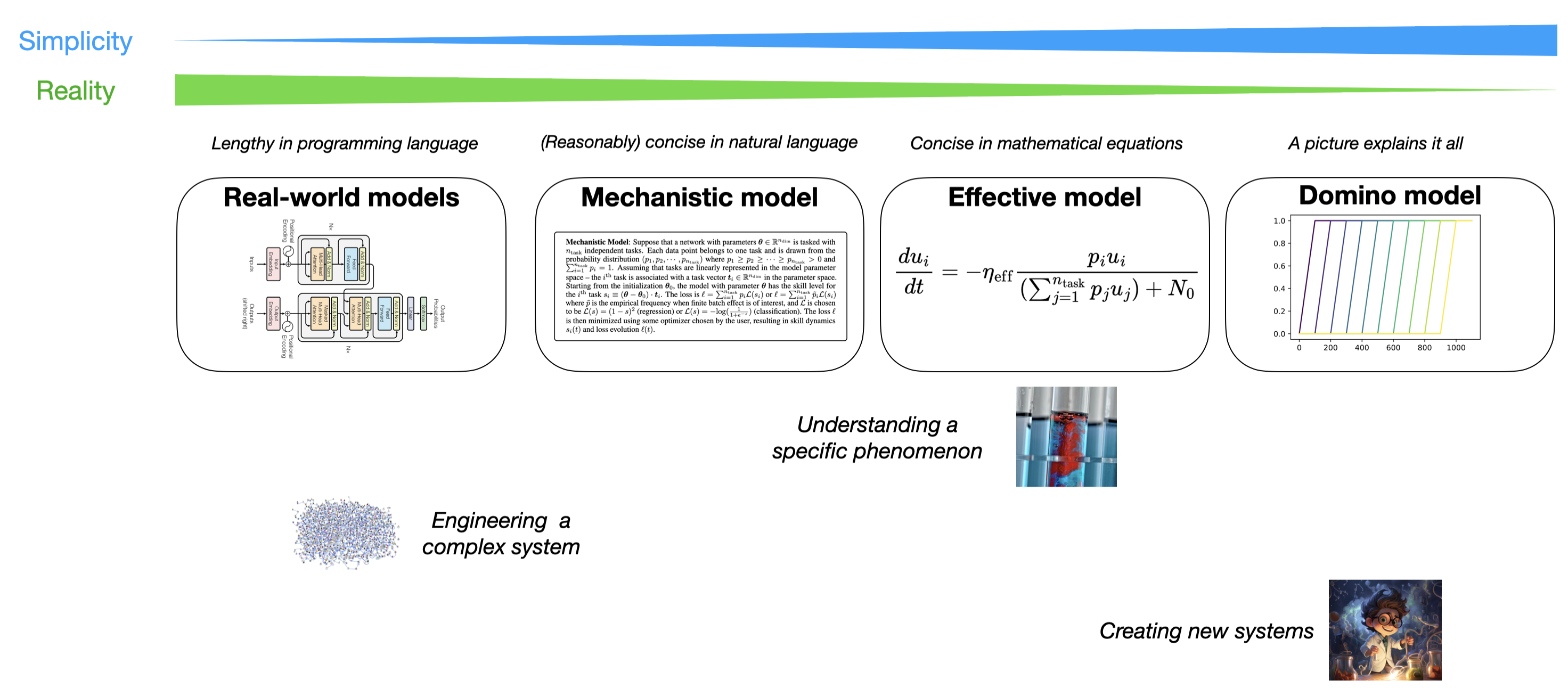}
    \caption{Instead of having a single model, it is better to have a spectrum of models. People can choose an appropriate model based on their goals.}
    \label{fig:conclusion}
\end{figure}

We have presented three simplified models for skill learning. Despite their simplicity and intuitiveness, they are able to shed light on many aspects of skill learning -- Domino effect (sequential learning of skills), non-monotonic dynamics, neural scaling laws, as well as how skill dynamics are affected by task dependency, modularity, and optimizers.

{\bf Limitations} However, these simplified models must miss some (perhaps quite important) aspects.
\begin{itemize}
    \item A notable limitation of our models is that they do not describe overfitting -- our analysis only applies to training data, which is valid for test data only when the generalization gap is small. However, the generalization gap is not described by our models.
    \item Another limitation is convexity -- the loss function in the Geometry model is a weighted average of convex functions (both MSE and cross-entropy are convex functions), which may not be expressive enough to represent non-convex landscapes. 
    \item  Lastly, we did not provide a recipe to map concrete data/tasks into our abstract notation of ``skills''. In particular, we have assumed that each data point belongs to one and only one skill, which may not be the complete story in practice. We may be able to integrate the skill framework by Arora et al.~\cite{arora2023theory} to obtain more fine-grained models to better bridge theory and practice.  
\end{itemize}

{\bf When are simplified models useful?} Despite the limitations, our models can be useful in the following situations (After all, as George Box puts it, ``All models are wrong, but some are useful''):

\begin{itemize}
    \item Simplifying mental pictures to enable intuitive thinking. Neural networks are complex systems with many degrees of freedom. Humans are not good at reasoning with or controlling systems with too many degrees of freedom. Our models, especially the Resource model that allows a resource interpretation, much simplifies the model of neural networks, facilitating understanding and making the creation of research ideas more intuitive. For example, our models can be used to study modularity (in Section~\ref{sec:modularity}), without going into the details of how modular networks are built. We also propose the reweighting technique (in Section~\ref{subsec:weighting}), inspired by the insights from the models.
    \item The Geometry model could be a good starting point (benchmark) for optimizers. To demonstrate the effectiveness of optimizers, people either use too simple examples (toy low-dimensional functions) or too complex examples (training language models). There is no such benchmark in between, that is simple enough to make intuition with, cheap enough that all researchers afford to study, but also complex enough to induce rich dynamics in real-world training. Our analysis in Section~\ref{subsec:optimizer} showcases how to use the Geometry model to understand the behavior of new optimizers, revealing their strengths and limitations which can transfer to large language models. 
\end{itemize}

From a pedagogical perspective, we want a spectrum of models that trade-off between simplicity and reality (shown in Figure~\ref{fig:conclusion}). There is no silver bullet model because one needs to choose the most appropriate model based on their goals. When we care about (a) understanding specific phenomena, we may want to adopt Einstein's advice ``make the model as simple as possible, but not simpler''. This is the strategy we adopted when attempting to understand the Domino effect. When we care about (b) creating new systems that are quite different from existing ones, it is helpful to have a simple and intuitive model, and only flesh out the details later. This is the strategy we adopted when we studied how modularity can speed up learning in Section~\ref{sec:modularity}. However, when we care about (c) engineering a complex system, it might be good to deal with the real-world system directly, although simplified models may provide some conceptual insight into how to better engineer the complex system. 

\section*{Acknowledgement}

Z.L., E.J.M. and M.T. are supported by IAIFI through NSF grant PHY-2019786. Z.L. is supported through the Google PhD Fellowship. E.J.M. is supported through the NSF GRFP
(Grant No. 2141064). We would like to thank Eric Alm, Mehran Kardar, and PLS AI Journal Club members for their helpful suggestions. 

\bibliographystyle{unsrt}
\bibliography{ref}

\begin{thebibliography}{10}

\bibitem{arora2023theory}
Sanjeev Arora and Anirudh Goyal.
\newblock A theory for emergence of complex skills in language models.
\newblock {\em arXiv preprint arXiv:2307.15936}, 2023.

\bibitem{chen2024skill}
Mayee Chen, Nicholas Roberts, Kush Bhatia, Jue Wang, Ce~Zhang, Frederic Sala, and Christopher R{\'e}.
\newblock Skill-it! a data-driven skills framework for understanding and training language models.
\newblock {\em Advances in Neural Information Processing Systems}, 36, 2024.

\bibitem{michaud2024quantization}
Eric Michaud, Ziming Liu, Uzay Girit, and Max Tegmark.
\newblock The quantization model of neural scaling.
\newblock {\em Advances in Neural Information Processing Systems}, 36, 2024.

\bibitem{wei2022emergent}
Jason Wei, Yi~Tay, Rishi Bommasani, Colin Raffel, Barret Zoph, Sebastian Borgeaud, Dani Yogatama, Maarten Bosma, Denny Zhou, Donald Metzler, et~al.
\newblock Emergent abilities of large language models.
\newblock {\em arXiv preprint arXiv:2206.07682}, 2022.

\bibitem{ortiz2024task}
Guillermo Ortiz-Jimenez, Alessandro Favero, and Pascal Frossard.
\newblock Task arithmetic in the tangent space: Improved editing of pre-trained models.
\newblock {\em Advances in Neural Information Processing Systems}, 36, 2024.

\bibitem{ilharco2022editing}
Gabriel Ilharco, Marco~Tulio Ribeiro, Mitchell Wortsman, Suchin Gururangan, Ludwig Schmidt, Hannaneh Hajishirzi, and Ali Farhadi.
\newblock Editing models with task arithmetic.
\newblock {\em arXiv preprint arXiv:2212.04089}, 2022.

\bibitem{todd2023function}
Eric Todd, Millicent~L Li, Arnab~Sen Sharma, Aaron Mueller, Byron~C Wallace, and David Bau.
\newblock Function vectors in large language models.
\newblock {\em arXiv preprint arXiv:2310.15213}, 2023.

\bibitem{kaplan2020scaling}
Jared Kaplan, Sam McCandlish, Tom Henighan, Tom~B Brown, Benjamin Chess, Rewon Child, Scott Gray, Alec Radford, Jeffrey Wu, and Dario Amodei.
\newblock Scaling laws for neural language models.
\newblock {\em arXiv preprint arXiv:2001.08361}, 2020.

\bibitem{hoffmann2022training}
Jordan Hoffmann, Sebastian Borgeaud, Arthur Mensch, Elena Buchatskaya, Trevor Cai, Eliza Rutherford, Diego de~Las Casas, Lisa~Anne Hendricks, Johannes Welbl, Aidan Clark, et~al.
\newblock Training compute-optimal large language models.
\newblock {\em arXiv preprint arXiv:2203.15556}, 2022.

\bibitem{besiroglu2024chinchilla}
Tamay Besiroglu, Ege Erdil, Matthew Barnett, and Josh You.
\newblock Chinchilla scaling: A replication attempt.
\newblock {\em arXiv preprint arXiv:2404.10102}, 2024.

\bibitem{power2022grokking}
Alethea Power, Yuri Burda, Harri Edwards, Igor Babuschkin, and Vedant Misra.
\newblock Grokking: Generalization beyond overfitting on small algorithmic datasets.
\newblock {\em arXiv preprint arXiv:2201.02177}, 2022.

\bibitem{lin2017focal}
T~Lin.
\newblock Focal loss for dense object detection.
\newblock {\em arXiv preprint arXiv:1708.02002}, 2017.

\bibitem{lin2024rho}
Zhenghao Lin, Zhibin Gou, Yeyun Gong, Xiao Liu, Yelong Shen, Ruochen Xu, Chen Lin, Yujiu Yang, Jian Jiao, Nan Duan, et~al.
\newblock Rho-1: Not all tokens are what you need.
\newblock {\em arXiv preprint arXiv:2404.07965}, 2024.

\bibitem{chen2024symbolic}
Xiangning Chen, Chen Liang, Da~Huang, Esteban Real, Kaiyuan Wang, Hieu Pham, Xuanyi Dong, Thang Luong, Cho-Jui Hsieh, Yifeng Lu, et~al.
\newblock Symbolic discovery of optimization algorithms.
\newblock {\em Advances in neural information processing systems}, 36, 2024.

\bibitem{pagliardini2024ademamix}
Matteo Pagliardini, Pierre Ablin, and David Grangier.
\newblock The ademamix optimizer: Better, faster, older.
\newblock {\em arXiv preprint arXiv:2409.03137}, 2024.

\bibitem{liu2023sophia}
Hong Liu, Zhiyuan Li, David Hall, Percy Liang, and Tengyu Ma.
\newblock Sophia: A scalable stochastic second-order optimizer for language model pre-training.
\newblock {\em arXiv preprint arXiv:2305.14342}, 2023.

\bibitem{vyas2024soap}
Nikhil Vyas, Depen Morwani, Rosie Zhao, Itai Shapira, David Brandfonbrener, Lucas Janson, and Sham Kakade.
\newblock Soap: Improving and stabilizing shampoo using adam.
\newblock {\em arXiv preprint arXiv:2409.11321}, 2024.

\bibitem{yuan2024mars}
Huizhuo Yuan, Yifeng Liu, Shuang Wu, Xun Zhou, and Quanquan Gu.
\newblock Mars: Unleashing the power of variance reduction for training large models.
\newblock {\em arXiv preprint arXiv:2411.10438}, 2024.

\bibitem{jordan2024muon}
Keller Jordan, Yuchen Jin, Vlado Boza, You Jiacheng, Franz Cecista, Laker Newhouse, and Jeremy Bernstein.
\newblock Muon: An optimizer for hidden layers in neural networks, 2024.

\bibitem{kingma2014adam}
Diederik~P Kingma.
\newblock Adam: A method for stochastic optimization.
\newblock {\em arXiv preprint arXiv:1412.6980}, 2014.

\bibitem{shalev2017failures}
Shai Shalev-Shwartz, Ohad Shamir, and Shaked Shammah.
\newblock Failures of gradient-based deep learning.
\newblock In {\em International Conference on Machine Learning}, pages 3067--3075. PMLR, 2017.

\bibitem{pfeiffer2023modular}
Jonas Pfeiffer, Sebastian Ruder, Ivan Vuli{\'c}, and Edoardo~Maria Ponti.
\newblock Modular deep learning.
\newblock {\em arXiv preprint arXiv:2302.11529}, 2023.

\bibitem{andreas2016neural}
Jacob Andreas, Marcus Rohrbach, Trevor Darrell, and Dan Klein.
\newblock Neural module networks.
\newblock In {\em Proceedings of the IEEE conference on computer vision and pattern recognition}, pages 39--48, 2016.

\bibitem{alet2018modular}
Ferran Alet, Tom{\'a}s Lozano-P{\'e}rez, and Leslie~P Kaelbling.
\newblock Modular meta-learning.
\newblock In {\em Conference on robot learning}, pages 856--868. PMLR, 2018.

\bibitem{masoudnia2014mixture}
Saeed Masoudnia and Reza Ebrahimpour.
\newblock Mixture of experts: a literature survey.
\newblock {\em Artificial Intelligence Review}, 42:275--293, 2014.

\bibitem{cai2024survey}
Weilin Cai, Juyong Jiang, Fan Wang, Jing Tang, Sunghun Kim, and Jiayi Huang.
\newblock A survey on mixture of experts.
\newblock {\em Authorea Preprints}, 2024.

\bibitem{cloud2024gradient}
Alex Cloud, Jacob Goldman-Wetzler, Ev{\v{z}}en Wybitul, Joseph Miller, and Alexander~Matt Turner.
\newblock Gradient routing: Masking gradients to localize computation in neural networks.
\newblock {\em arXiv preprint arXiv:2410.04332}, 2024.

\bibitem{mohri2018foundations}
Mehryar Mohri.
\newblock Foundations of machine learning, 2018.

\bibitem{engel2001statistical}
Andreas Engel.
\newblock {\em Statistical mechanics of learning}.
\newblock Cambridge University Press, 2001.

\bibitem{bahri2020statistical}
Yasaman Bahri, Jonathan Kadmon, Jeffrey Pennington, Sam~S Schoenholz, Jascha Sohl-Dickstein, and Surya Ganguli.
\newblock Statistical mechanics of deep learning.
\newblock {\em Annual Review of Condensed Matter Physics}, 11(1):501--528, 2020.

\bibitem{jacot2018neural}
Arthur Jacot, Franck Gabriel, and Cl{\'e}ment Hongler.
\newblock Neural tangent kernel: Convergence and generalization in neural networks.
\newblock {\em Advances in neural information processing systems}, 31, 2018.

\bibitem{watanabe2007almost}
Sumio Watanabe.
\newblock Almost all learning machines are singular.
\newblock In {\em 2007 IEEE Symposium on Foundations of Computational Intelligence}, pages 383--388. IEEE, 2007.

\bibitem{wei2022deep}
Susan Wei, Daniel Murfet, Mingming Gong, Hui Li, Jesse Gell-Redman, and Thomas Quella.
\newblock Deep learning is singular, and that’s good.
\newblock {\em IEEE Transactions on Neural Networks and Learning Systems}, 34(12):10473--10486, 2022.

\bibitem{ren2024towards}
Qihan Ren, Yang Xu, Junpeng Zhang, Yue Xin, Dongrui Liu, and Quanshi Zhang.
\newblock Towards the dynamics of a dnn learning symbolic interactions.
\newblock {\em arXiv preprint arXiv:2407.19198}, 2024.

\bibitem{hendel2023context}
Roee Hendel, Mor Geva, and Amir Globerson.
\newblock In-context learning creates task vectors.
\newblock {\em arXiv preprint arXiv:2310.15916}, 2023.

\bibitem{yang2025taskvectorsincontextlearning}
Liu Yang, Ziqian Lin, Kangwook Lee, Dimitris Papailiopoulos, and Robert Nowak.
\newblock Task vectors in in-context learning: Emergence, formation, and benefit, 2025.

\bibitem{olsson2022context}
Catherine Olsson, Nelson Elhage, Neel Nanda, Nicholas Joseph, Nova DasSarma, Tom Henighan, Ben Mann, Amanda Askell, Yuntao Bai, Anna Chen, et~al.
\newblock In-context learning and induction heads.
\newblock {\em arXiv preprint arXiv:2209.11895}, 2022.

\bibitem{liu2022towards}
Ziming Liu, Ouail Kitouni, Niklas~S Nolte, Eric Michaud, Max Tegmark, and Mike Williams.
\newblock Towards understanding grokking: An effective theory of representation learning.
\newblock {\em Advances in Neural Information Processing Systems}, 35:34651--34663, 2022.

\bibitem{liu2022omnigrok}
Ziming Liu, Eric~J Michaud, and Max Tegmark.
\newblock Omnigrok: Grokking beyond algorithmic data.
\newblock In {\em The Eleventh International Conference on Learning Representations}, 2022.

\bibitem{nanda2023progress}
Neel Nanda, Lawrence Chan, Tom Lieberum, Jess Smith, and Jacob Steinhardt.
\newblock Progress measures for grokking via mechanistic interpretability.
\newblock {\em arXiv preprint arXiv:2301.05217}, 2023.

\bibitem{abbe2023sgd}
Emmanuel Abbe, Enric~Boix Adsera, and Theodor Misiakiewicz.
\newblock Sgd learning on neural networks: leap complexity and saddle-to-saddle dynamics.
\newblock In {\em The Thirty Sixth Annual Conference on Learning Theory}, pages 2552--2623. PMLR, 2023.

\bibitem{zhong2024clock}
Ziqian Zhong, Ziming Liu, Max Tegmark, and Jacob Andreas.
\newblock The clock and the pizza: Two stories in mechanistic explanation of neural networks.
\newblock {\em Advances in Neural Information Processing Systems}, 36, 2024.

\bibitem{cui2024phase}
Hugo Cui, Freya Behrens, Florent Krzakala, and Lenka Zdeborov{\'a}.
\newblock A phase transition between positional and semantic learning in a solvable model of dot-product attention.
\newblock {\em arXiv preprint arXiv:2402.03902}, 2024.

\bibitem{baek2024geneft}
David~D Baek, Ziming Liu, and Max Tegmark.
\newblock Geneft: Understanding statics and dynamics of model generalization via effective theory.
\newblock {\em arXiv preprint arXiv:2402.05916}, 2024.

\bibitem{hayou2019impact}
Soufiane Hayou, Arnaud Doucet, and Judith Rousseau.
\newblock On the impact of the activation function on deep neural networks training.
\newblock In {\em International conference on machine learning}, pages 2672--2680. PMLR, 2019.

\bibitem{yang2020feature}
Greg Yang and Edward~J Hu.
\newblock Feature learning in infinite-width neural networks.
\newblock {\em arXiv preprint arXiv:2011.14522}, 2020.

\bibitem{schaeffer2024emergent}
Rylan Schaeffer, Brando Miranda, and Sanmi Koyejo.
\newblock Are emergent abilities of large language models a mirage?
\newblock {\em Advances in Neural Information Processing Systems}, 36, 2024.

\bibitem{sharma2020neural}
Utkarsh Sharma and Jared Kaplan.
\newblock A neural scaling law from the dimension of the data manifold.
\newblock {\em arXiv preprint arXiv:2004.10802}, 2020.

\bibitem{song2024resource}
Jinyeop Song, Ziming Liu, Max Tegmark, and Jeff Gore.
\newblock A resource model for neural scaling law.
\newblock {\em arXiv preprint arXiv:2402.05164}, 2024.

\bibitem{bahri2024explaining}
Yasaman Bahri, Ethan Dyer, Jared Kaplan, Jaehoon Lee, and Utkarsh Sharma.
\newblock Explaining neural scaling laws.
\newblock {\em Proceedings of the National Academy of Sciences}, 121(27):e2311878121, 2024.

\bibitem{maloney2022solvable}
Alexander Maloney, Daniel~A Roberts, and James Sully.
\newblock A solvable model of neural scaling laws.
\newblock {\em arXiv preprint arXiv:2210.16859}, 2022.

\bibitem{bordelon2024dynamical}
Blake Bordelon, Alexander Atanasov, and Cengiz Pehlevan.
\newblock A dynamical model of neural scaling laws.
\newblock {\em arXiv preprint arXiv:2402.01092}, 2024.

\bibitem{sorscher2022beyond}
Ben Sorscher, Robert Geirhos, Shashank Shekhar, Surya Ganguli, and Ari Morcos.
\newblock Beyond neural scaling laws: beating power law scaling via data pruning.
\newblock {\em Advances in Neural Information Processing Systems}, 35:19523--19536, 2022.

\bibitem{liu2024kan}
Ziming Liu, Yixuan Wang, Sachin Vaidya, Fabian Ruehle, James Halverson, Marin Solja{\v{c}}i{\'c}, Thomas~Y Hou, and Max Tegmark.
\newblock Kan: Kolmogorov-arnold networks.
\newblock {\em arXiv preprint arXiv:2404.19756}, 2024.

\bibitem{du2022glam}
Nan Du, Yanping Huang, Andrew~M Dai, Simon Tong, Dmitry Lepikhin, Yuanzhong Xu, Maxim Krikun, Yanqi Zhou, Adams~Wei Yu, Orhan Firat, et~al.
\newblock Glam: Efficient scaling of language models with mixture-of-experts.
\newblock In {\em International Conference on Machine Learning}, pages 5547--5569. PMLR, 2022.

\bibitem{kumar2024scaling}
Tanishq Kumar, Zachary Ankner, Benjamin~F Spector, Blake Bordelon, Niklas Muennighoff, Mansheej Paul, Cengiz Pehlevan, Christopher R{\'e}, and Aditi Raghunathan.
\newblock Scaling laws for precision.
\newblock {\em arXiv preprint arXiv:2411.04330}, 2024.

\bibitem{frantar2023scaling}
Elias Frantar, Carlos Riquelme, Neil Houlsby, Dan Alistarh, and Utku Evci.
\newblock Scaling laws for sparsely-connected foundation models.
\newblock {\em arXiv preprint arXiv:2309.08520}, 2023.

\bibitem{liang2024cautious}
Kaizhao Liang, Lizhang Chen, Bo~Liu, and Qiang Liu.
\newblock Cautious optimizers: Improving training with one line of code.
\newblock {\em arXiv preprint arXiv:2411.16085}, 2024.

\bibitem{large2024scalable}
Tim Large, Yang Liu, Minyoung Huh, Hyojin Bahng, Phillip Isola, and Jeremy Bernstein.
\newblock Scalable optimization in the modular norm.
\newblock {\em arXiv preprint arXiv:2405.14813}, 2024.

\bibitem{gupta2018shampoo}
Vineet Gupta, Tomer Koren, and Yoram Singer.
\newblock Shampoo: Preconditioned stochastic tensor optimization.
\newblock In {\em International Conference on Machine Learning}, pages 1842--1850. PMLR, 2018.

\end{thebibliography}

\newpage
\appendix
\section{Collapse of learning curves}\label{app:collase}

\begin{figure}[ht]
    \centering
    \includegraphics[width=0.8\linewidth]{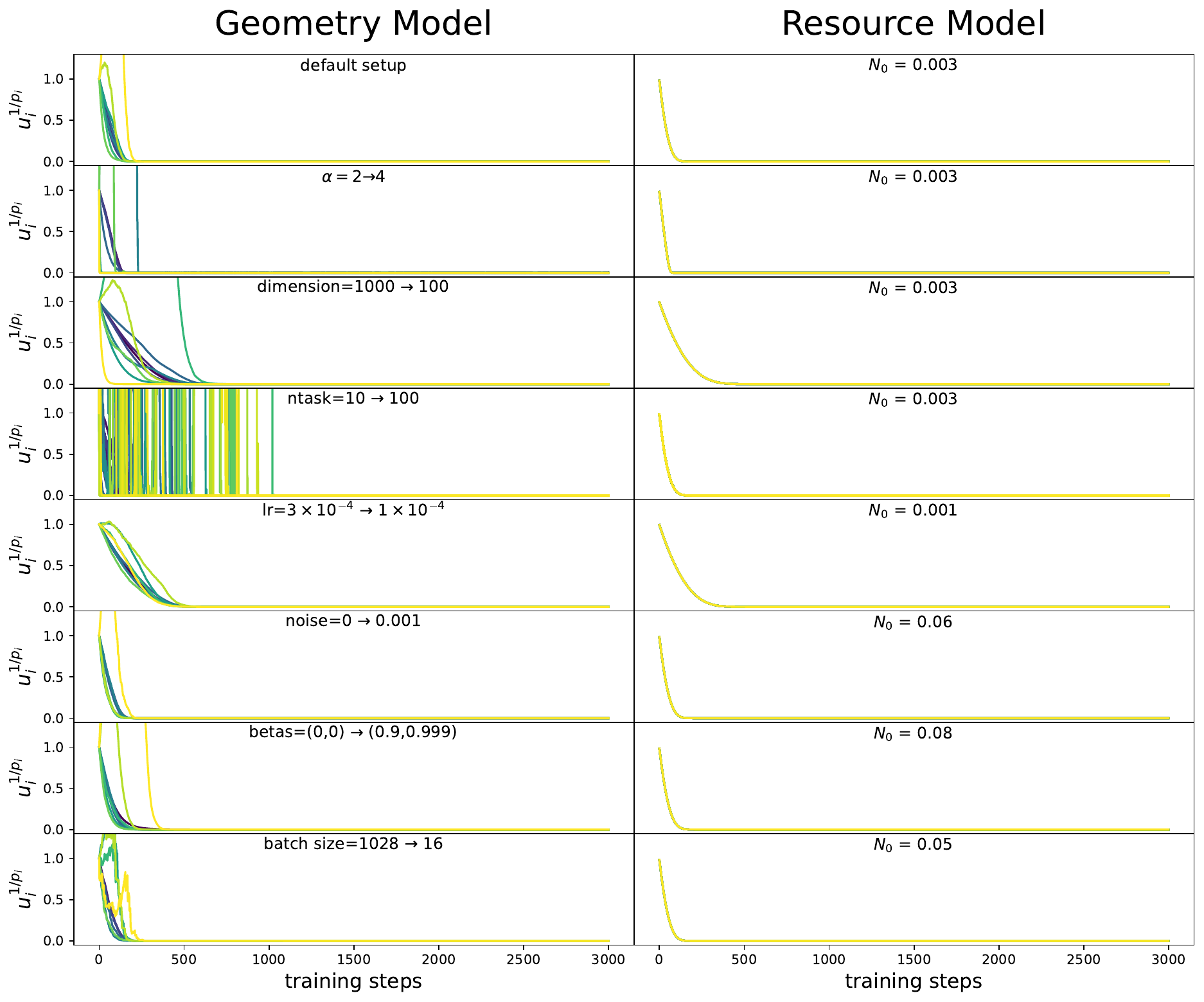}
    \caption{For the Resource model, we have proved that $u_i^{1/p_i}=C$ for all skills (learning curve collapse), which is verified in the right column. We find that the collapse of the learning curve approximately holds for the Geometry model as well (except for very infrequent skills, colored yellow. This is because when $p_i\ll 1$, exponentiating a number to the power of $1/p_i\gg 1$ can be very sensitive or unstable.).} 
    \label{fig:experiment_theory_compare_collapse}
\end{figure}

\end{document}